\documentclass[letterpaper, 10 pt, conference]{ieeeconf}
\IEEEoverridecommandlockouts    

\usepackage{multirow}
\usepackage{lipsum}
\usepackage{amsmath}
\usepackage{amssymb}
\usepackage[ruled,vlined]{algorithm2e}
\usepackage{graphicx}
\graphicspath{{./Figures/}}
\usepackage{xcolor}

\usepackage{comment}
\usepackage{booktabs}
\usepackage{tabularx}
\usepackage{subcaption}
\usepackage{float}
\usepackage{hyperref}
\usepackage{bbm}

\usepackage{xcolor}

\definecolor{RellisAsphalt}{RGB}{64, 64, 64}
\definecolor{RellisBarrier}{RGB}{41, 121, 255}
\definecolor{RellisBush}{RGB}{255, 153, 204}
\definecolor{RellisConcrete}{RGB}{170, 170, 170}
\definecolor{RellisFence}{RGB}{102, 0, 204}
\definecolor{RellisGrass}{RGB}{0, 102, 0}
\definecolor{RellisLog}{RGB}{102, 0, 0}
\definecolor{RellisMud}{RGB}{99, 66, 34}
\definecolor{RellisPole}{RGB}{0, 153, 153}
\definecolor{RellisPuddle}{RGB}{134, 255, 239}
\definecolor{RellisRubble}{RGB}{110, 22, 138}
\definecolor{RellisTree}{RGB}{0, 255, 0}
\definecolor{RellisWater}{RGB}{0, 128, 255}

\definecolor{WildBush}{RGB}{230, 25, 75}
\definecolor{WildDirt}{RGB}{60, 180, 75}
\definecolor{WildFence}{RGB}{0, 128, 128}
\definecolor{WildGrass}{RGB}{128, 128, 128}
\definecolor{WildGravel}{RGB}{145, 30, 180}
\definecolor{WildLog}{RGB}{128, 128, 0}
\definecolor{WildMud}{RGB}{255, 225, 25}
\definecolor{WildRock}{RGB}{170, 255, 195}
\definecolor{WildStructure}{RGB}{170, 110, 40}
\definecolor{WildTreeFoliage}{RGB}{210, 245, 60}
\definecolor{WildTreeTrunk}{RGB}{240, 50, 230}

\title{\LARGE \bf Improving Zero-shot 3D Semantic Mapping in Offroad with Geometric Priors}

\author{
 	\parbox{\textwidth}{%
 		\centering
 		Eerik Alamikkotervo$^{1}$, Risto Ojala$^{1}$
 	}%
 	\thanks{$^{1}$Energy and Mechanical Engineering, Aalto University, Espoo, Finland
 		{\tt\small eerik.alamikkotervo@aalto.fi, risto.ojala@aalto.fi}}%
}

\begin{document}
	
\maketitle
\thispagestyle{empty}
\pagestyle{empty}

\begin{abstract}
    Semantic 3D mapping is a powerful representation that enables platform agnostic mobile robot navigation in challenging offroad conditions. Current vision foundation models can accurately predict semantics zero-shot on structured scenes but performance is compromised in natural environments. On the other hand, 3D point clouds encode valuable information for discriminating semantics in scenarios where vision foundation models fail. In this work, we propose fitting a sphericity-elevation distribution for each semantic class to act as geometric anchor. Only the points with the most confident vision foundation model predictions are considered when fitting to produce high quality geometric prior distributions. The geometric prior is fused with the vision foundation model prediction to produce the final semantic mapping. The proposed prior improves SAM3 baseline by +6.3 in mIoU and +8.9 in accuracy on the RELLIS-3D dataset, and by +1.7 in mIoU and +2.1 in accuracy on the WildScenes dataset. The source code will be made public upon paper acceptance. 
\end{abstract}

\section{Introduction}
Semantic understanding of the environment is crucial for autonomous navigation in offroad conditions. Most offroad research concentrates on traversability estimation that reduces the rich geometry and semantic information to single dimension \cite{jung2024v,kim2026geo3tp}. These methods commonly classify any elevated 3D points as non-traversable and cannot accurately reason about the passability of different types of vegetation. Semantic 3D perception would provide more flexible foundation for offroad navigation that can be easily adapted to different scenarios and platforms. 

While the current foundation models for general purpose segmentation \cite{carion2026sam,ghiasi2022scaling} achieve impressive accuracy on structured environments their performance is significantly lower on unstructured natural scenes. Separating different kinds of vegetation and terrain from images is one of the major challenges as there are no clear visual boundaries. At the same time correct classification of different vegetation and terrain types is critical for succesful navigation decisions as they cover significant portion of natural scenes and have different traversability properties. 

Different terrain and vegetation types typically have distinctive geometric features that can help image based foundation models discriminate between them. We propose utilizing elevation and sphericity computed from lidar point clouds for this task. Elevation separates the terrain from vegetation and obstacles above it. Sphericity discriminates between different terrain and vegetation types based on their local shape. We fit an elevation-sphericity histogram for each semantic class based on the most confident detections from the vision foundation model to act as an geometric anchor. We then place prior for each point by comparing its elevation and sphericity to the fitted histograms. This prior is fused with the vision foundation model prediction to produce the final estimate. The proposed method yields significant improvements over vision foundation model baselines when evaluated with two real-world offroad datasets: Rellis3d \cite{jiang2021rellis} and WildScenes \cite{vidanapathirana2025wildscenes}. The main contributions are:

\begin{itemize}
    \item Novel elevation-sphericity histogram based prior for 3D semantic segmentation in offroad environments
    \item Demonstration of significant improvements in both mIoU and accuracy with real-world offroad datasets.
    \item Open-source implementation to support future development
\end{itemize}

\begin{figure}
    \centering
    \includegraphics[width=1.0\linewidth]{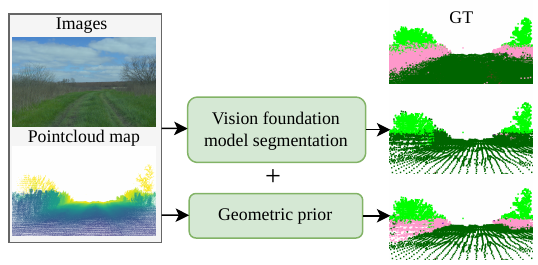}
    \caption{The proposed method improves semantic mapping by fusing the vision foundation model prediction with a geometric prior computed from the pointcloud.}
    \label{fig:intro}
\end{figure}

\section{Related work}

\subsection{Vision foundation models for zero-shot semantic segmentation}

Historically, segmentation models have been trained on task- and domain-specific datasets to achieve reliable performance within that domain.
The current trend is to pretrain large vision models on massive, diverse datasets using self-supervised, weakly supervised, or multimodal objectives, and then adapt them to a variety of downstream tasks and domains with minimal additional training. These models are referred to as vision foundation models. Thanks to extensive pre-training many vision foundation models can be used in zero-shot manner, meaning they require no training in the target domain to perform inference. Variety of vision foundation models have been proposed in recent years. DINO \cite{oquab2023dinov2} and EVA \cite{fang2023eva} are popular examples of feature extraction backbones that can be fine tuned for variety of downstream tasks. SAM \cite{kirillov2023segment} and Grounded SAM \cite{ren2024grounded} are designed for the segmentation task. Large vision language models like Gemini \cite{team2023gemini} and Qwen VL \cite{bai2025qwen3} support image inputs but they are trained for higher level image understanding and reasoning not pixel level segmentation task.

Another recent trend has been the deveploment of open vocabulary models to allow segmenting images with arbitary prompts. SAM3 \cite{carion2026sam} added support to query masks with short open vocabulary concenpts. DINO.txt \cite{jose2025dinov2} aligns dense DINO features with text prompt encodings to to allow open vocabulary segmentation. OpenSeg \cite{ghiasi2022scaling} employs a separate mask proposal network that generates class agnostic masks proposals that are then classified based on the feature similarity with the encoded text prompts.

\subsection{Zero-shot segmentation in 3D}

Traditional supervised 3D segmentation networks are trained to produce semantic labels from the pointcloud with manually labeled dataset. For zero-shot methods a common approach is to predict features or segmentation masks from the image with vision foundation models and lift them to 3D with known calibration \cite{yang2023sam3d, jatavallabhula2023conceptfusion}. Image based foundation models provide better zero-shot open vocabulary performance thanks to better training data availability, most importantly image-caption pairs. Image foundation models can also serve as supervision to distill 3D prediction models \cite{peng2023openscene,zhu2024open} enabling zero-shot open-vocabulary segmentation from 3D inputs alone, while also allowing image and 3D modalities to be fused for improved performance. However, the performance gains are limited as the 3D model relies on the image model for supervision leading to correlated outputs. 

\subsection{Zero-shot segmentation in offroad}

Offroad conditions pose spesific challenges to zero-shot segmentation in both 3D and image modalities. Images lack easily identifiable salient objects and most foundation model training data is from structured scenes which leads to decreased performance in offroad environments. In 3D, points don't necessarily form dense easily identifiable clusters as in the common 3D segmentation benchmarks like ScanNet \cite{dai2017scannet} and  Matterport3D \cite{chang2017matterport3d}. Instead the point density is singificantly lower and shapes more irregular. 

Very limited research has been conducted to solve zero-shot semantic segmentation in offroad conditions. Instead, most works concentrate on estimating just the traversability in a self-supervised manner \cite{jung2024v,kim2026geo3tp}. However, such approaches can't effectively handle scenarios where the platform needs to navigate over variety of terrain, obstacles and vegetation with different passability. 3D semantic maps are also inheritly platform agnostic unlike most self-supervised traversability estimation methods. Traversability is property tied to a specific platform while 3D semantic mapping is an objective description of the environment.     

Best of our knowledge OTAS \cite{schwaiger2025otas} is the only work aiming to specifically address the issue of zero-shot semantic segmentation in offroad conditions. They demonstrate that dense CLIP \cite{radford2021learning} features commonly used for open vocabulary segmentation are noisy in the offroad scenario. To solve this problem they extract dense DINOv2 features that are visually more robust and cluster them to form mask proposals. These mask proposals are then classified using their average CLIP feature. In the 3D case the point coordinates are concatenated with the features before clustering to encourage spatial coherency.      

\subsection{Geometric features in offroad robotics}

Geometric features have been long used for evaluating the environment properties in offroad conditions. Common features are step-height, roughness and slope \cite{borges2022survey}. These features work well in solid terrain, but not when passing through vegetation is required. High vegetation typically has high roughness and step height leading to classification as obstacle. Utilization of linearity, planarity and sphericity features is one proposed solution to this problem \cite{ruetz2022forest, bellone2017learning}. Vegetation typically has high sphericity, whereas terrain and structures exhibit high planarity, and thin obstacles high linearity.    

\subsection{Research Gap}

We improve zero-shot 3D semantic mapping by utilizing geometric priors based on elevation and sphericity extracted from lidar point clouds. Refrence elevation-sphericity histogram is fitted for each class utilizing the points with most confident vision foundation model predictions. The fitting process is lightweight and the histogram describes the geometrical separation of different classes in human understandable form. In existing works, OTAS utilizes 3D coordinates to improve spatial segmentation coherency but ignores other cues available from geometry \cite{schwaiger2025otas}. On the other hand, distilling 3D networks with image based foundation model supervision requires environment specific neural network training that is time and computation heavy.  \cite{peng2023openscene,zhu2024open}. 

\section{Methods}
\label{sec:methods}

\begin{figure}
    \centering
    \includegraphics[width=0.9\linewidth]{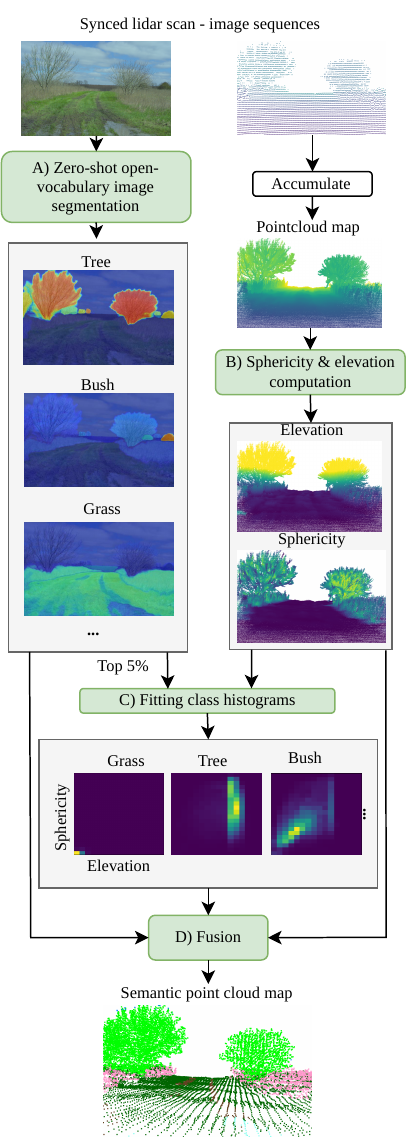}
    \caption{Geometric prior computation and utilization illustrated. The most confident image segmentation predictions are used to fit the sphericity-elevation distributions for each class, which are then used to assign prior weights for each point in the fusion phase.}
    \label{fig:placeholder}
\end{figure}

Given a synced lidar scan-image sequence our method generates a semantic pointcloud map of the sequence. The method is divided to four parts. First confidence score maps are generated for each class using an open vocabulary vision foundation model (Section \ref{sec:image_seg}). Then sphericity and elevation are computed for the 3D point cloud map accumulated from the posed scans (Section \ref{sec:sph_elev}). Then sphericity-elevation histogram is fitted using the points with most confident vision foundation model predictions (Section \ref{sec:fit}). Finally, the geometric prior is queried from the fitted histograms and fused with the vision foundation model confidence maps to produce the final segmentation output (Section \ref{sec:fusion}). 

\subsection{Zero-shot open vocabulary image segmentation}
\label{sec:image_seg}

We employ an open vocabulary vision foundation model for the image segmentation task. Given an open vocabulary prompt the model outputs a score for each pixel or image patch describing the correspondence with that prompt. To define a class wise score distribution each image is prompted with all the class names in the dataset yielding confidence score map $\mathbf{S}_{\text{img}}^t \in \mathbb{R}^{H \times W \times C}$ at each timestep $t$, where $H,W$ is the image size and $C$ number of semantic classes. 
We then project the lidar scan from timestep $t$ to the image frame based on known intrinsic and extrinsic calibration and retrieve the semantic information from the corresponding pixels yielding pointwise confidence score map $\mathbf{S}_{\text{scan}}^t  \in \mathbb{R}^{n \times C}$, where $n$ is the number of points in the scan. Points that are outside the camera field of view are discarded. Points are accumulated across the sequence based on known poses to form a confidence score point cloud map $\mathbf{S}_{\text{map}}  \in \mathbb{R}^{N \times C}$, where $N$ is total number of scan points in the sequence. A corresponding 3D cloud map $\textbf{P}_{\text{map}}  \in \mathbb{R}^{N \times 3}$ is also formed.   

\subsection{Sphericity and elevation computation}
\label{sec:sph_elev}
\begin{figure}
    \centering
    \includegraphics[width=1.0\linewidth]{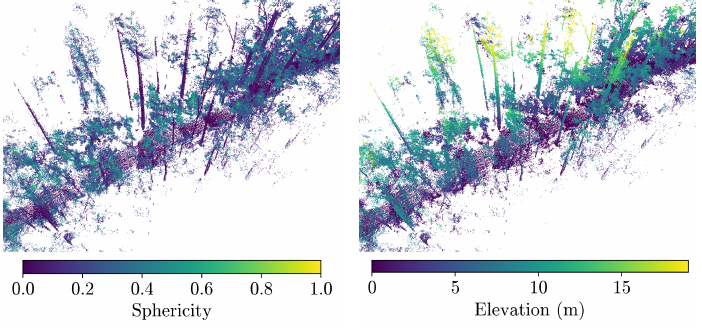}
    \caption{Sphericity and elevation visualized. Tree trunks and even ground have low sphericity and tree foliage and vegetation high sphericity.}
    \label{fig:elev-sphe}
\end{figure}

Sphericity and elevation are both computed over the accumulated point cloud $\textbf{P}_{\text{map}}$. The computations are performed on 0.1 m voxel resolution to save computation and results are upscaled back to the full resolution yielding elevation map $\textbf{P}_{e} \in \mathbb{R}^{N}$ and sphericity map $\mathbf{P}_{s} \in \mathbb{R}^{N}$. Elevation and sphericity maps are visualized in Figure \ref{fig:elev-sphe}. 

The relative elevation $e_{\mathbf{p}}$ of a point $\mathbf{p}$ is defined as the difference between its height $z_\mathbf{p}$ and the minimum height within its local neighborhood
\begin{equation}
    e_\mathbf{p} = z_\mathbf{p} - \min_{\mathbf{q} \in \mathcal{N}^{2D}_r(\mathbf{p})} z_\mathbf{q},
\end{equation}
where $\mathcal{N}_r^{2\text{D}}(\mathbf{p})$ denotes the local neighborhood comprising all points within a horizontal radius $r = 0.5\text{ m}$ of $\mathbf{p}$. The orientation of the z-coordinate is aligned with the normal vector of the global point cloud's principal plane, determined via principal component analysis. 

To compute sphericity, we first construct the 3D covariance matrix $\mathbf{\Sigma}_\mathbf{p} \in \mathbb{R}^{3 \times 3}$ over the local 3D neighborhood $\mathcal{N}^{3D}_r(\mathbf{p})$, which includes all points within a sphere of radius $r = 0.5\text{ m}$ centered at $\mathbf{p}$:
\begin{equation}
\mathbf{\Sigma}_\mathbf{p} = \frac{1}{k} \sum_{\mathbf{q} \in \mathcal{N}^{3D}_r(\mathbf{p})} (\mathbf{q} - \mathbf{\bar{q}})(\mathbf{q} - \mathbf{\bar{q}})^T,
\end{equation}
where $k$ is the number of points in the neighborhood and $\mathbf{\bar{q}}$ is their centroid. Let $\lambda_1 \ge \lambda_2 \ge \lambda_3 \ge 0$ denote the eigenvalues of $\mathbf{\Sigma}_\mathbf{p}$. The sphericity $s_\mathbf{p}$ of point $\mathbf{p}$ is then defined as:
\begin{equation}
    s_\mathbf{p} = \frac{\lambda_3}{\lambda_1}.
\end{equation}

\subsection{Fitting class histograms}
\label{sec:fit}

\begin{figure}
    \centering
    \includegraphics[width=1.0\linewidth]{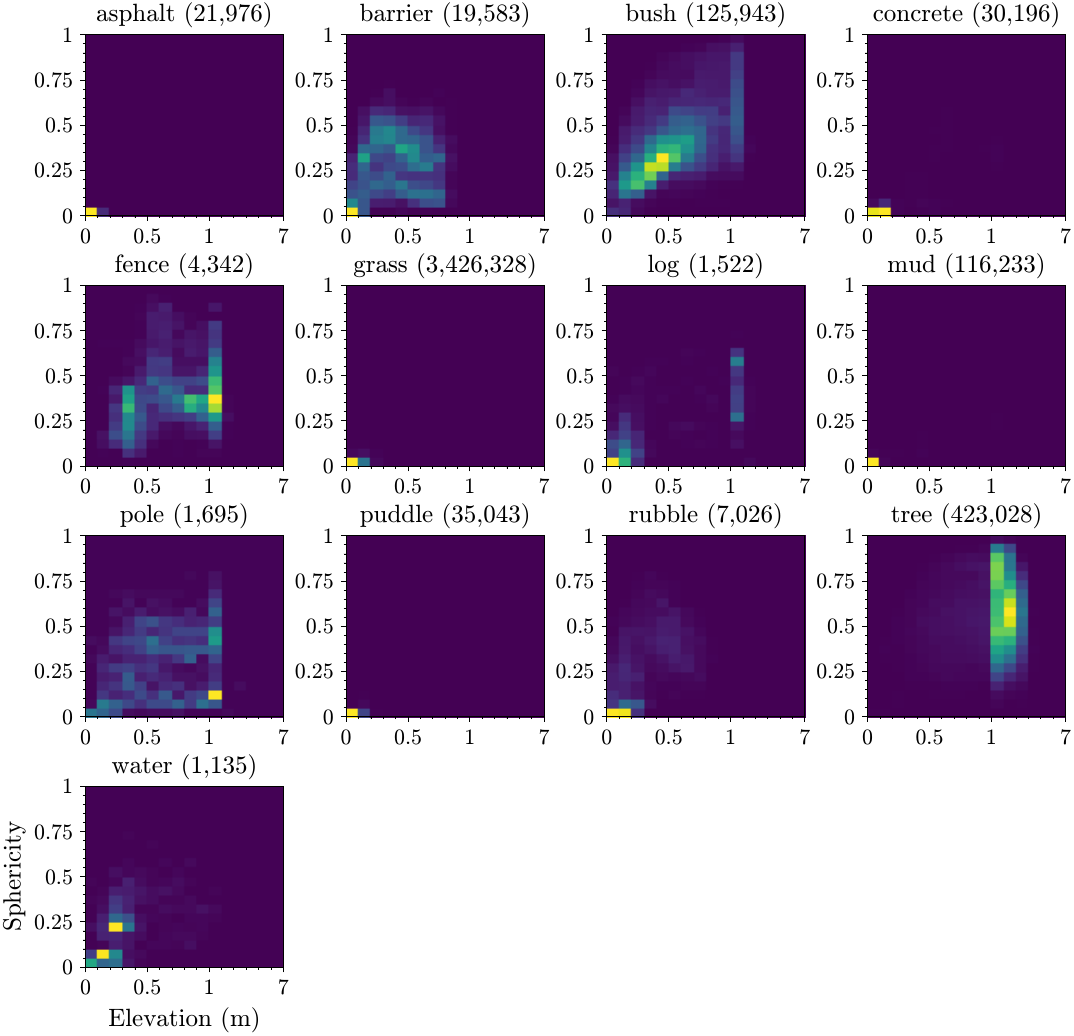}
    \caption{Elevation-sphericity histograms fitted on the top 5 \% SAM3 detections on Rellis3D. Number of samples used for fitting each class in parentheses.}
    \label{fig:rellis_hist}
\end{figure}

\begin{figure}
    \centering
    \includegraphics[width=1.0\linewidth]{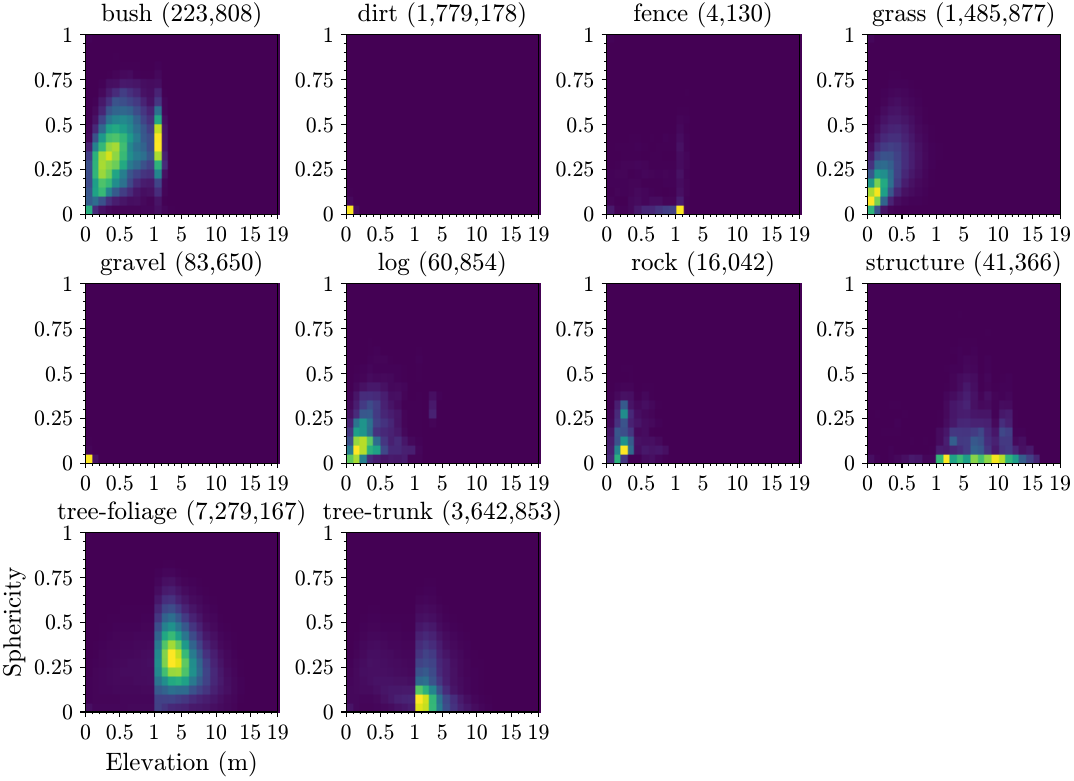}
    \caption{Elevation-sphericity histograms fitted on the top 5 \% SAM3 detections on WildScenes. Mud excluded due to insufficient number of points. Number of samples used for fitting each class in parentheses.}
    \label{fig:wildscenes_hist}
\end{figure}

We model the geometric properties of each class with 2D histograms that can accurately capture arbitrary shapes unlike parametric distributions. For elevation, we define non-uniform bins $E_i$ utilizing a 0.1 m bin spacing for values between 0--1 m, and a 1.0 m bin spacing for values exceeding 1 m. For sphericity, we define uniform bins $S_j$ with a 0.05 spacing across the entire $[0, 1]$ range. To fit the distribution for a given class $c$, we utilize only the most confident predictions to avoid incorporating noisy samples. Let $\mathcal{P}_{5\%}^c$ denote this high-confidence subset, consisting of the top 5\% most confident points among all points where class $c$ is the highest-scoring prediction from the vision foundation model. Let $i$ be the elevation bin index and $j$ be the sphericity bin index. The unnormalized occurrence count $h^c(i, j)$ for class $c$ is computed by accumulating the points in $\mathcal{P}_{5\%}^c$ that fall into the respective bins:
\begin{equation}
h^c(i, j) = \alpha + \sum_{\mathbf{p} \in \mathcal{P}_{5\%}^c} \mathbbm{1} \Big( e_\mathbf{p} \in E_i \land s_\mathbf{p} \in S_j \Big),
\end{equation}
where $\mathbbm{1}(\cdot)$ is the indicator function, and $\alpha = 1$ is the initial count added to every bin to avoid zero counts that veto out that class. Finally, we normalize the accumulated counts to sum to one across the histogram:
\begin{equation} 
h^c_\text{norm}(i,j) = \frac{h^c(i,j)}{\sum_{u} \sum_{v} h^c(u,v)},
\end{equation}
where $u$ and $v$ iterate over all elevation and sphericity bins, respectively. The fitted histograms for each class in the Rellis3D and WildScenes datasets are presented in Figures \ref{fig:rellis_hist} and \ref{fig:wildscenes_hist}.  

\subsection{Fusion}
\label{sec:fusion}

We can construct a geometric prior matrix $\mathbf{S}_{\text{prior}} \in \mathbb{R}^{N \times C}$ for $\mathbf{S}_{\text{map}}$ by querying the normalized class histogram $h_\text{norm}^c$ at each points $\mathbf{p}$ and each class $c$: 
\begin{equation}
\mathbf{S}_{\text{prior}}[x_\mathbf{p}, c] = h_\text{norm}^c \big( i(e_\mathbf{p}), j(s_\mathbf{p}) \big),
\end{equation}
where $i(e_\mathbf{p})$ and $j(s_\mathbf{p})$ are the bin indeces of the point's elevation and sphericity respectively and $x_\mathbf{p}$ is the points $\mathbf{p}$ index. 

If both distributions $\textbf{S}_{\text{map}}$ and $\textbf{S}_{\text{prior}}$ were calibrated likelihoods, their direct product would yield a standard Bayesian fused estimate. Because strict calibration cannot be assumed, we introduce a scalar smoothing exponent $w$ to control the weighting between the geometry and vision modalities. The fused score matrix $\mathbf{S}_{\text{fused}} \in \mathbb{R}^{N \times C}$ is computed via the element-wise product:
\begin{equation}
\label{eq:fusion}
\mathbf{S}_{\text{fused}} = \mathbf{S}_{\text{prior}}^{\circ w} \odot \mathbf{S}_{\text{map}},
\end{equation}
where $\odot$ denotes element-wise multiplication, and $(\cdot)^{\circ w}$ denotes element-wise exponentiation. Setting $w=1$ corresponds to standard Bayesian fusion, whereas $w=0$ ignores the geometric prior entirely. 

We apply the prior only to the set of well-sampled classes $\mathcal{U}$ whose histograms $h^c$ exceed 1,000 point samples. While rare classes $c \notin \mathcal{U}$ retain their original scores from $\mathbf{S}_{\text{map}}$, classes in $\mathcal{U}$ are fused via Equation~\ref{eq:fusion} and rescaled to preserve their original collective score sum in $\mathbf{S}_{\text{map}}$. Consequently, the prior redistributes confidence exclusively among classes in $\mathcal{U}$ and rest remain unchanged. 

The final predicted class for each point is extracted by taking the row-wise argument maximum, yielding a class assignment vector $\mathbf{c}^* \in \mathbb{Z}^N$:
\begin{equation}
\mathbf{c}^* = \underset{c \in \mathcal{C}}{\arg\max} \; \mathbf{S}_{\text{fused}}.
\end{equation}
Finally, the semantically labeled point cloud is downsampled to a 0.2 m voxel resolution, where each voxel is assigned the class that receives the majority vote of all points in the voxel.

\subsection{Validation}

We evaluate the proposed method using three zero-shot, open-vocabulary image segmentation models: DINOv2.txt, SAM 3, and OpenSeg. Given a text prompt, SAM 3 outputs a set of mask proposals, each paired with a mask score in the range $[0, 1]$ representing the model's confidence in that mask, as well as a pixel-wise score in $[0, 1]$ indicating the confidence that each pixel belongs to that mask. We define the final per-pixel confidence score as the product of the global mask score and the pixel-wise mask score. DINOv2.txt employs a ViT-L/14 backbone with registers alongside its corresponding vision head and text encoder. OpenSeg uses an EfficientNet-B7 backbone paired with CLIP ViT-L/14@336px text embeddings, following OpenScene \cite{peng2023openscene}. For both DINOv2.txt and OpenSeg, we scale the raw cosine similarities between the text prompt embeddings and visual embeddings by a temperature of $T = 100$ before applying a softmax function, consistent with the CLIP training recipe.All models operate at their respective recommended input resolutions: SAM 3 inputs are stretched directly to $1008 \times 1008\text{ px}$, OpenSeg inputs are resized to a $640\text{ px}$ long side while preserving the aspect ratio, with the short side padded to $640\text{ px}$ and DINOv2.txt inputs are resized to a $224\text{ px}$ short side while preserving the aspect ratio. 

Additionally, we compare against OTAS, which is specifically designed for zero-shot 3D semantic segmentation in unstructured environments. For each feature patch, we assign the median depth of all LiDAR scan points projecting into that patch. Following the authors' recommended configuration, we use the spatial variant of OTAS with 24 clusters, 24 components, a DINOv2 ViT-S/14 backbone with registers, and a $224\text{ px}$ input resolution, setting the voxel size to $0.2\text{ m}$ to match our method's resolution. The reconstructed map is queried with every dataset class prompt, and each 3D point is assigned the class yielding the highest similarity score. To keep GPU memory consumption manageable, the map is processed in chunks of 500 samples.

\subsection{Data}

We evaluate our method using the WildScenes \cite{vidanapathirana2025wildscenes} and Rellis3D \cite{jiang2021rellis} datasets. Both include image-lidar scan pairs with known calibration and ground truth for pointcloud semantics. We only consider the lidar scan points that are inside the camera field of view as all evaluated methods require image data in addition to the 3D points. Lidar points further than 25 m are filtered out as alignment errors increase and sensor resolution decreases with longer distances.   

WildScenes is collected on Australian forest trails surrounded by high tree canopies. We exclude ambigous classes: \textit{other-terrain}, \textit{other-object} and \textit{unlabelled}. We also exclude \textit{sky} and \textit{water} as these labels are not present in the 3D data.  

RELLIS3D is collected on rural fields and trails in Texas dominated by grass, bushes and smaller trees. We exclude ambigous classes \textit{void} and \textit{object}. We exclude \textit{sky} as it is not present in the 3D data and \textit{person} and \textit{vehicle} as they are dynamic. We also exclude \textit{building} and \textit{dirt} as no points belonging to these classes are retained with the camera fov filtering with 25 m max distance used here. The poses were computed with kiss-icp \cite{vizzo2023kiss}.   

\section{Results}
\label{sec:results}

Accuracy, mIoU and class wise IoU are reported for each model on Rellis3D dataset (Table \ref{tab:preds_rellis3d}) and WildScenes dataset (Table \ref{tab:preds_wildscenes}). The geometry prior improves performance for all benchmarked zero-shot segmentation models.  
SAM3 is the best performing baseline model achieving 28.1 mIoU and 73.9 accuracy on Rellis3d and 47.9 mIoU and 80.5 accuracy on WildScenes. With the geometry prior SAM3 achieves +6.3 mIoU and +8.9 accuracy increase on Rellis3d and +1.7 mIoU and +2.1 accuracy increase on WildScenes. 
On Rellis3d the classes with highest gains are Bush (+44.2), Grass (+8.8), Puddle (+5.2) and Water (+28.7) and on WildScenes Bush (+3.8), Dirt (+7.9) and Grass (+9.2). OTAS often confuses similar classes, like dirt and gravel or tree-foliage and tree-trunk and achieves slightly lower accuracy than the DINOv2.txt and OpenSeg baselines. Qualitative examples for OTAS, SAM3 and SAM3 with geometric priors are presented in Tables \ref{fig:qualitative} and \ref{fig:qualitative2}. 

For further analysis, we ablate different smoothing factors and the performance of only sphericity or elevation based prior histograms (Table \ref{tab:ablations}). With both datasets the joint elevation-sphericity prior outperforms either component alone. Optimal smoothing factor depends on the quality of the image predictions as it controls how much weight the geometry is given over the image predictions. Smoothing factor 0.25 was found to provide the best compromise across both datasets and is used with all reported results here. In most cases, any choice of the smoothing factor provided significant improvements over the baseline.

\begin{table*}[]
    \centering
    \caption{Semantic mapping performance on Rellis3d dataset. Highest values \bf{bolded}}
    \begin{tabular}{|l | c | c | c | c | c | c | c | c | c | c | c | c | c | c | c |}
        \hline
        \textbf{Method} & \textbf{mIoU} & \textbf{Acc} & \rotatebox{90}{\textcolor{RellisAsphalt}{\rule{2ex}{2ex}} \textbf{Asphalt}} & \rotatebox{90}{\textcolor{RellisBarrier}{\rule{2ex}{2ex}} \textbf{Barrier}} & \rotatebox{90}{\textcolor{RellisBush}{\rule{2ex}{2ex}} \textbf{Bush}} & \rotatebox{90}{\textcolor{RellisConcrete}{\rule{2ex}{2ex}} \textbf{Concrete}} & \rotatebox{90}{\textcolor{RellisFence}{\rule{2ex}{2ex}} \textbf{Fence}} & \rotatebox{90}{\textcolor{RellisGrass}{\rule{2ex}{2ex}} \textbf{Grass}} & \rotatebox{90}{\textcolor{RellisLog}{\rule{2ex}{2ex}} \textbf{Log}} & \rotatebox{90}{\textcolor{RellisMud}{\rule{2ex}{2ex}} \textbf{Mud}} & \rotatebox{90}{\textcolor{RellisPole}{\rule{2ex}{2ex}} \textbf{Pole}} & \rotatebox{90}{\textcolor{RellisPuddle}{\rule{2ex}{2ex}} \textbf{Puddle}} & \rotatebox{90}{\textcolor{RellisRubble}{\rule{2ex}{2ex}} \textbf{Rubble}} & \rotatebox{90}{\textcolor{RellisTree}{\rule{2ex}{2ex}} \textbf{Tree}} & \rotatebox{90}{\textcolor{RellisWater}{\rule{2ex}{2ex}} \textbf{Water}} \\
        \hline
        OTAS & 20.1 & 69.6 & 0.0 & 5.2 & 9.6 & 36.6 & 5.4 & 68.7 & 0.1 & 21.9 & 0.0 & 20.8 & 22.6 & \bf{70.8} & 0.0 \\
        \hline
        SAM 3 & 28.1 & 73.9 & 0.0 & \bf{68.7} & 7.2 & 42.7 & 5.8 & 73.4 & 12.1 & 30.2 & 0.3 & 28.5 & 27.1 & 68.8 & 0.0 \\
         + OURS & \bf{34.4} & \bf{82.8} & 0.0 & \bf{68.7} & \bf{49.4} & 42.2 & 4.6 & \bf{82.2} & \bf{12.6} & \bf{30.6} & 0.2 & \bf{33.7} & 26.9 & 67.5 & \bf{28.7} \\
        \hline
        DINOv2.txt & 21.0 & 65.5 & 0.0 & 18.7 & 41.0 & 43.2 & 3.1 & 65.5 & 9.7 & 16.9 & 0.7 & 18.3 & 37.0 & 19.2 & 0.0 \\
        + OURS & 22.5 & 68.0 & 0.0 & 18.3 & 46.5 & \bf{44.4} & 3.3 & 67.4 & 9.1 & 17.2 & 0.6 & 18.6 & \bf{37.2} & 30.3 & 0.0 \\
        \hline
        OpenSeg & 21.5 & 75.1 & 0.0 & 41.1 & 41.7 & 0.0 & 22.8 & 77.3 & 0.0 & 14.7 & \bf{9.3} & 22.4 & 25.3 & 24.9 & 0.0 \\
         + OURS & 23.1 & 77.4 & 0.0 & 50.1 & 47.4 & 0.0 & \bf{23.1} & 79.3 & 0.0 & 15.5 & 5.9 & 21.7 & 25.6 & 32.3 & 0.0 \\
        \hline
    \end{tabular}
    \label{tab:preds_rellis3d}
\end{table*}

\begin{table*}[]
    \centering
    \caption{Semantic mapping performance on WildScenes dataset. Highest values \bf{bolded}}
    \begin{tabular}{|l | c | c | c | c | c | c | c | c | c | c | c | c | c |}
        \hline
        \textbf{Method} & \textbf{mIoU} & \textbf{Acc} & \rotatebox{90}{\textcolor{WildBush}{\rule{2ex}{2ex}} \textbf{Bush}} & \rotatebox{90}{\textcolor{WildDirt}{\rule{2ex}{2ex}} \textbf{Dirt}} & \rotatebox{90}{\textcolor{WildFence}{\rule{2ex}{2ex}} \textbf{Fence}} & \rotatebox{90}{\textcolor{WildGrass}{\rule{2ex}{2ex}} \textbf{Grass}} & \rotatebox{90}{\textcolor{WildGravel}{\rule{2ex}{2ex}} \textbf{Gravel}} & \rotatebox{90}{\textcolor{WildLog}{\rule{2ex}{2ex}} \textbf{Log}} & \rotatebox{90}{\textcolor{WildMud}{\rule{2ex}{2ex}} \textbf{Mud}} & \rotatebox{90}{\textcolor{WildRock}{\rule{2ex}{2ex}} \textbf{Rock}} & \rotatebox{90}{\textcolor{WildStructure}{\rule{2ex}{2ex}} \textbf{Structure}} & \rotatebox{90}{\textcolor{WildTreeFoliage}{\rule{2ex}{2ex}} \textbf{Tree-Foliage}} & \rotatebox{90}{\textcolor{WildTreeTrunk}{\rule{2ex}{2ex}} \textbf{Tree-Trunk}} \\
        \hline
        OTAS & 14.5 & 22.1 & 3.4 & 15.3 & 7.5 & 40.9 & 5.5 & 0.0 & 0.0 & 0.1 & 69.0 & 0.4 & 16.9 \\
        \hline
        SAM 3 & 47.9 & 80.5 & 13.4 & 69.8 & 48.3 & 57.0 & 50.7 & \textbf{35.8} & 0.0 & \textbf{42.7} & 82.3 & 78.5 & 48.1 \\
         + OURS & \textbf{49.6} & \textbf{82.6} & \textbf{17.2} & \textbf{77.7} & 44.0 & \textbf{67.8} & \textbf{50.8} & 34.9 & 0.0 & 40.3 & \textbf{83.9} & \textbf{79.8} & \textbf{49.1} \\
        \hline
        DINOv2.txt & 21.3 & 24.0 & 1.2 & 23.3 & 7.3 & 17.7 & 7.5 & 14.3 & 1.5 & 31.1 & 73.8 & 17.0 & 39.1 \\
         + OURS & 23.4 & 33.2 & 1.4 & 27.1 & 6.5 & 22.6 & 8.0 & 14.3 & 1.3 & 31.6 & 74.1 & 29.3 & 41.4 \\
        \hline
        OpenSeg & 27.0 & 73.9 & 14.9 & 56.6 & \textbf{49.2} & 17.1 & 18.4 & 2.3 & \textbf{9.0} & 0.5 & 51.7 & 77.4 & 0.2 \\
         + OURS & 27.9 & 74.4 & 14.2 & 58.5 & 48.9 & 19.4 & 18.4 & 1.2 & 8.4 & 0.6 & 59.0 & 78.5 & 0.4 \\
        \hline
    \end{tabular}
    \label{tab:preds_wildscenes}
\end{table*}

\begin{table*}[]
    \centering
    \caption{Examples of semantic pointcloud map predictions from ego view.}
    \begin{tabularx}{\textwidth}{*{5}{>{\centering\arraybackslash}X}}
        \toprule
        \textbf{Image} & \textbf{Ground Truth} & \textbf{OTAS} & \textbf{SAM3} & \textbf{SAM3+OURS} \\
        \midrule
        
        \multicolumn{5}{c}{\textbf{RELLIS-3D}} \\
        \midrule
        
        \includegraphics[width=0.95\linewidth]{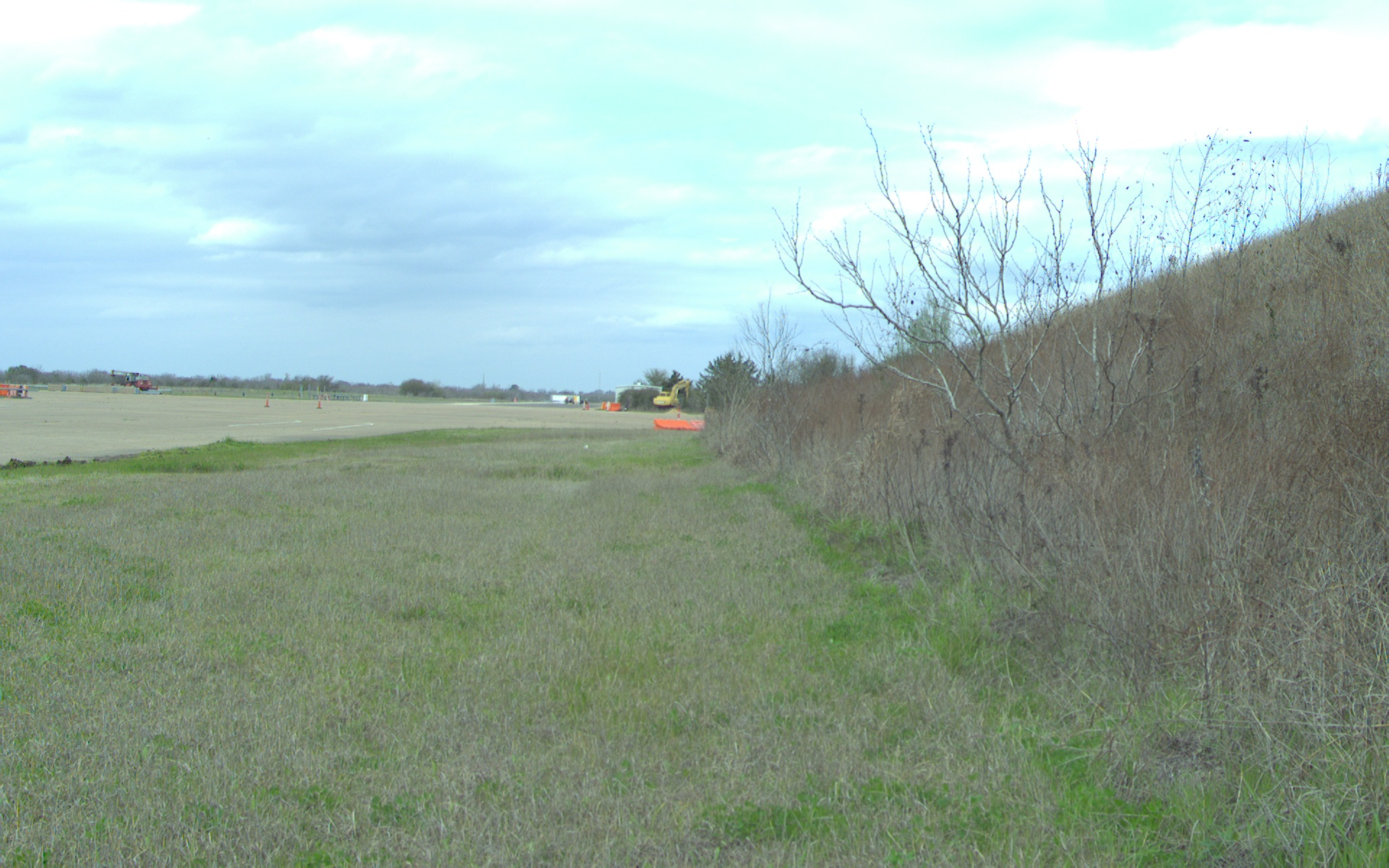} & 
        \includegraphics[width=0.95\linewidth]{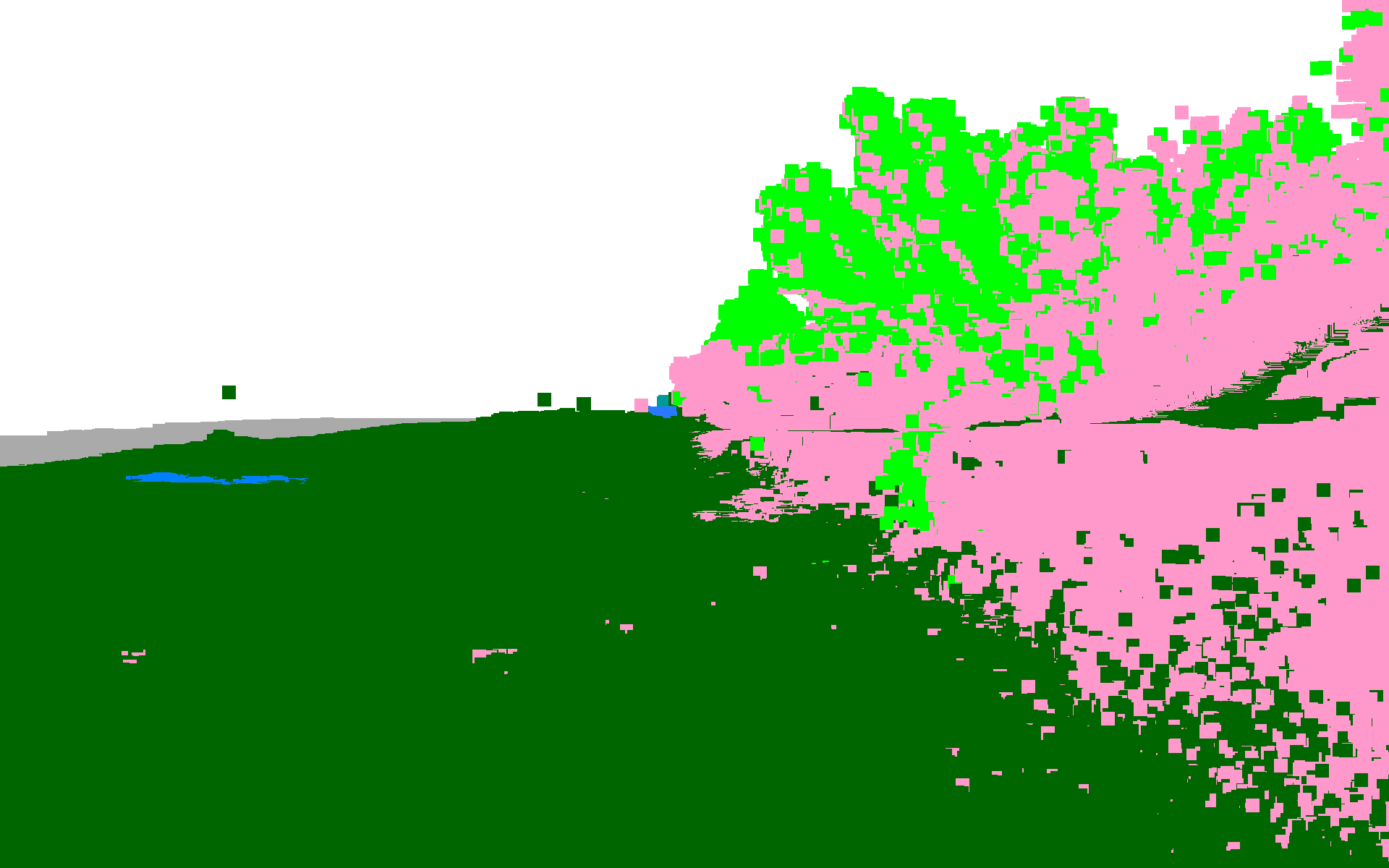} & 
        \includegraphics[width=0.95\linewidth]{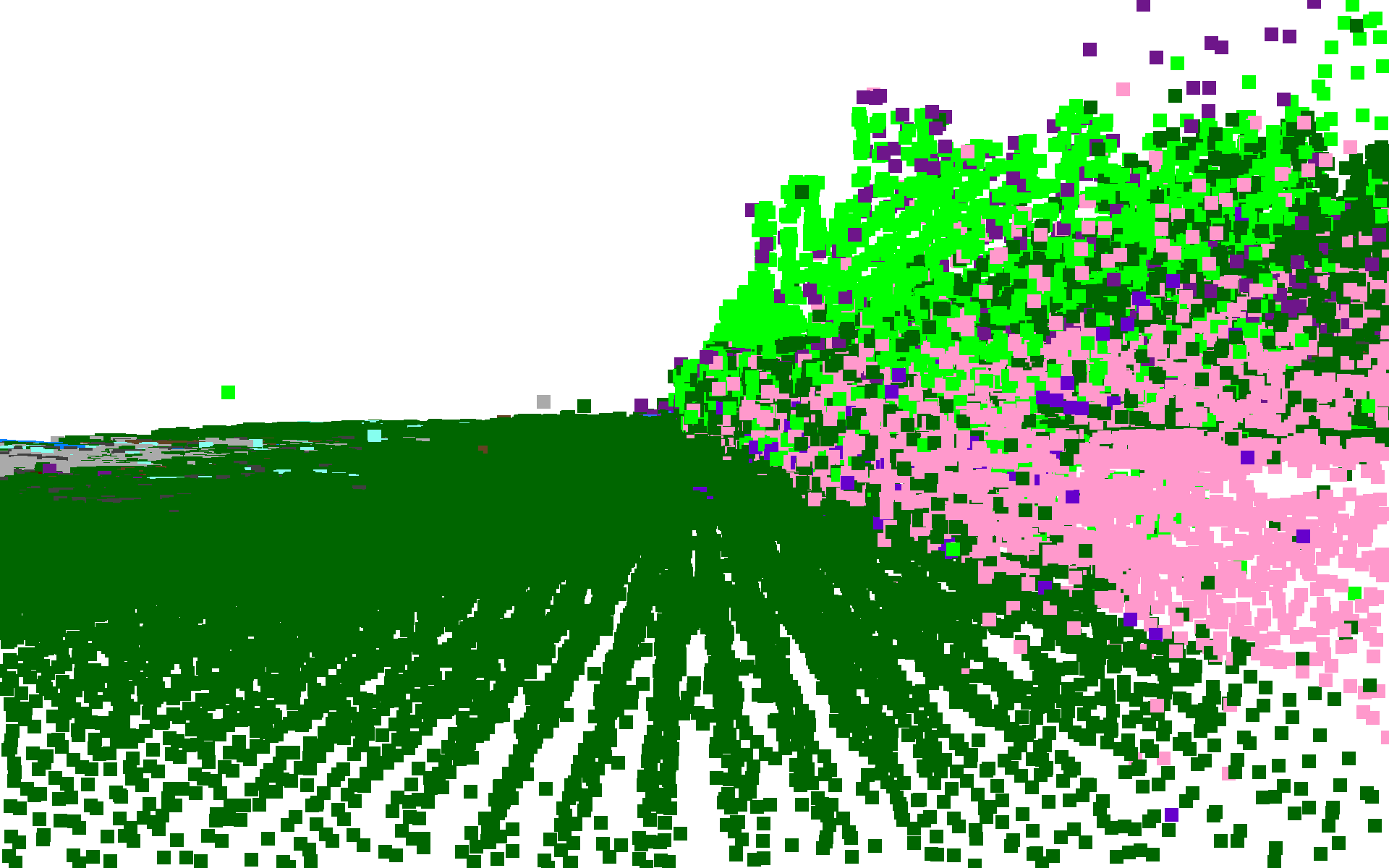} & 
        \includegraphics[width=0.95\linewidth]{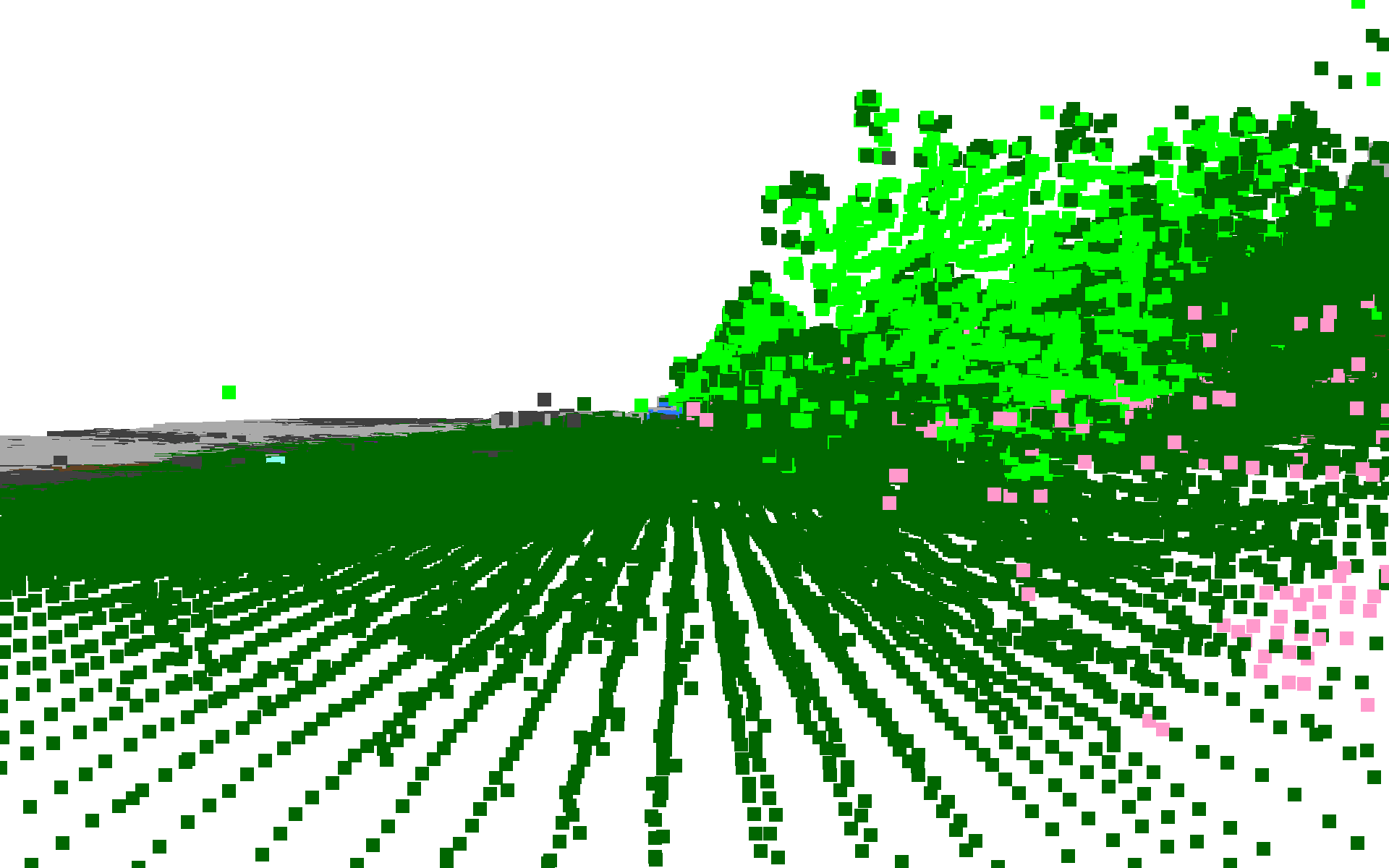} & 
        \includegraphics[width=0.95\linewidth]{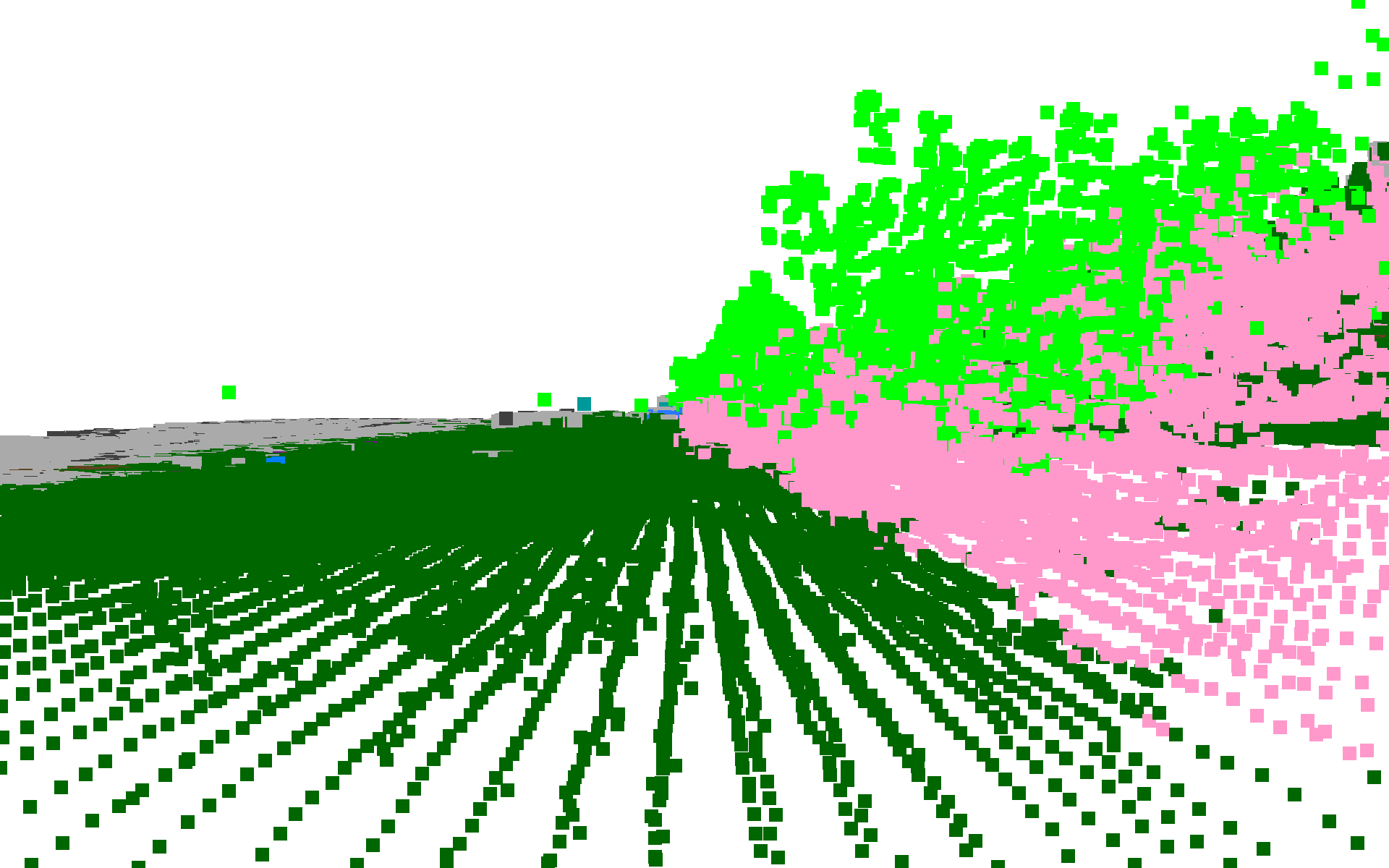} \\[0.2cm]

        \includegraphics[width=0.95\linewidth]{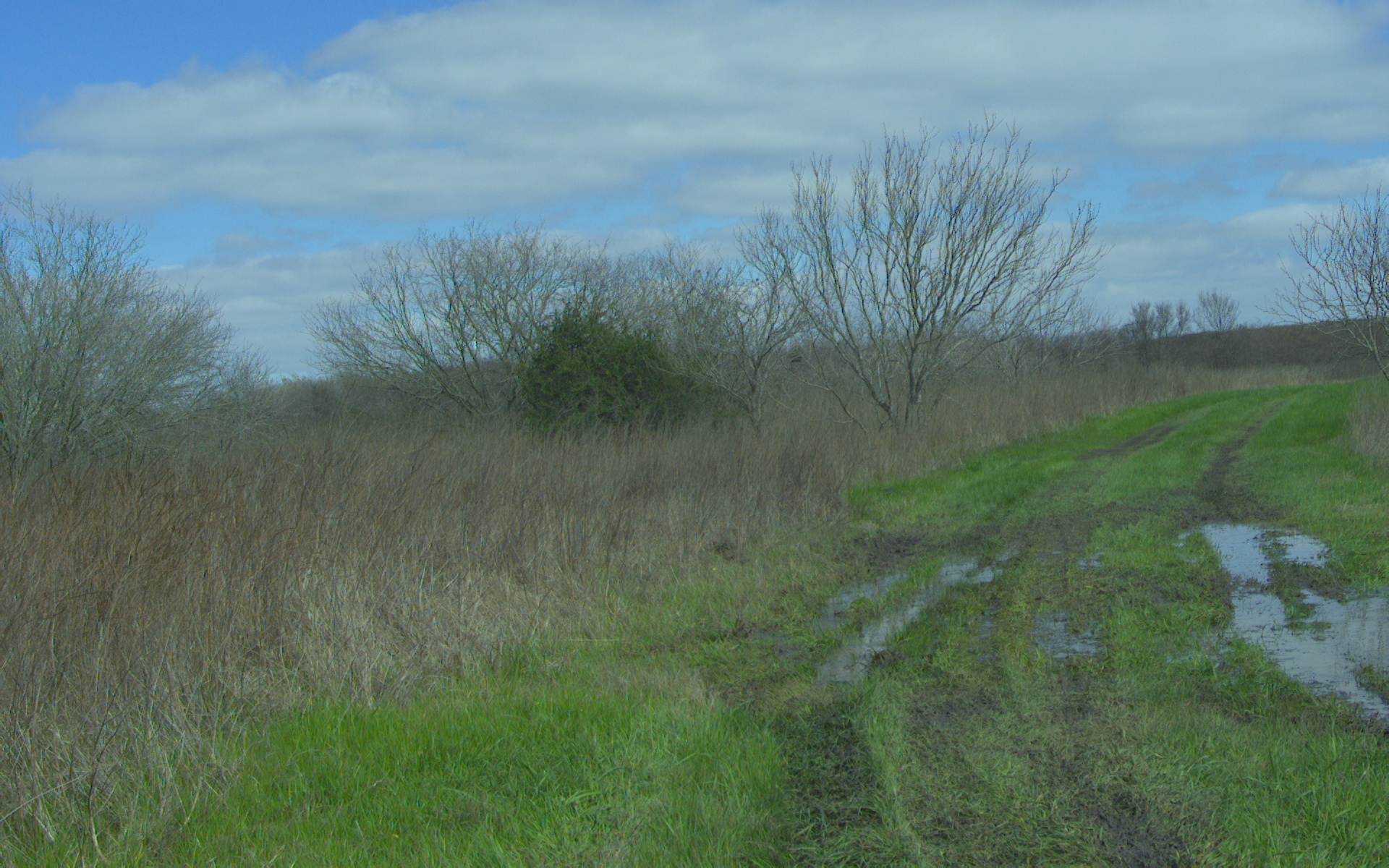} & 
        \includegraphics[width=0.95\linewidth]{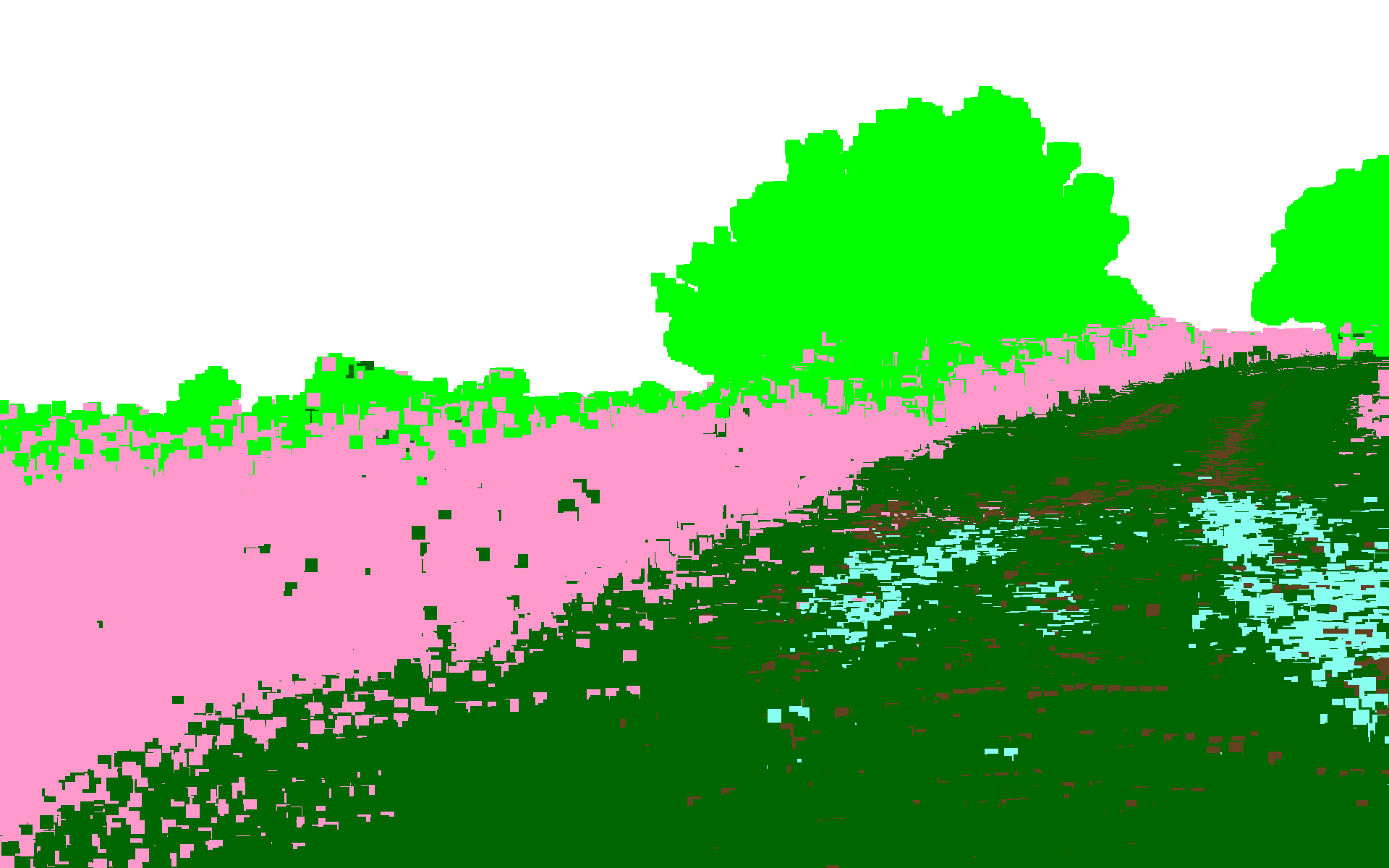} & 
        \includegraphics[width=0.95\linewidth]{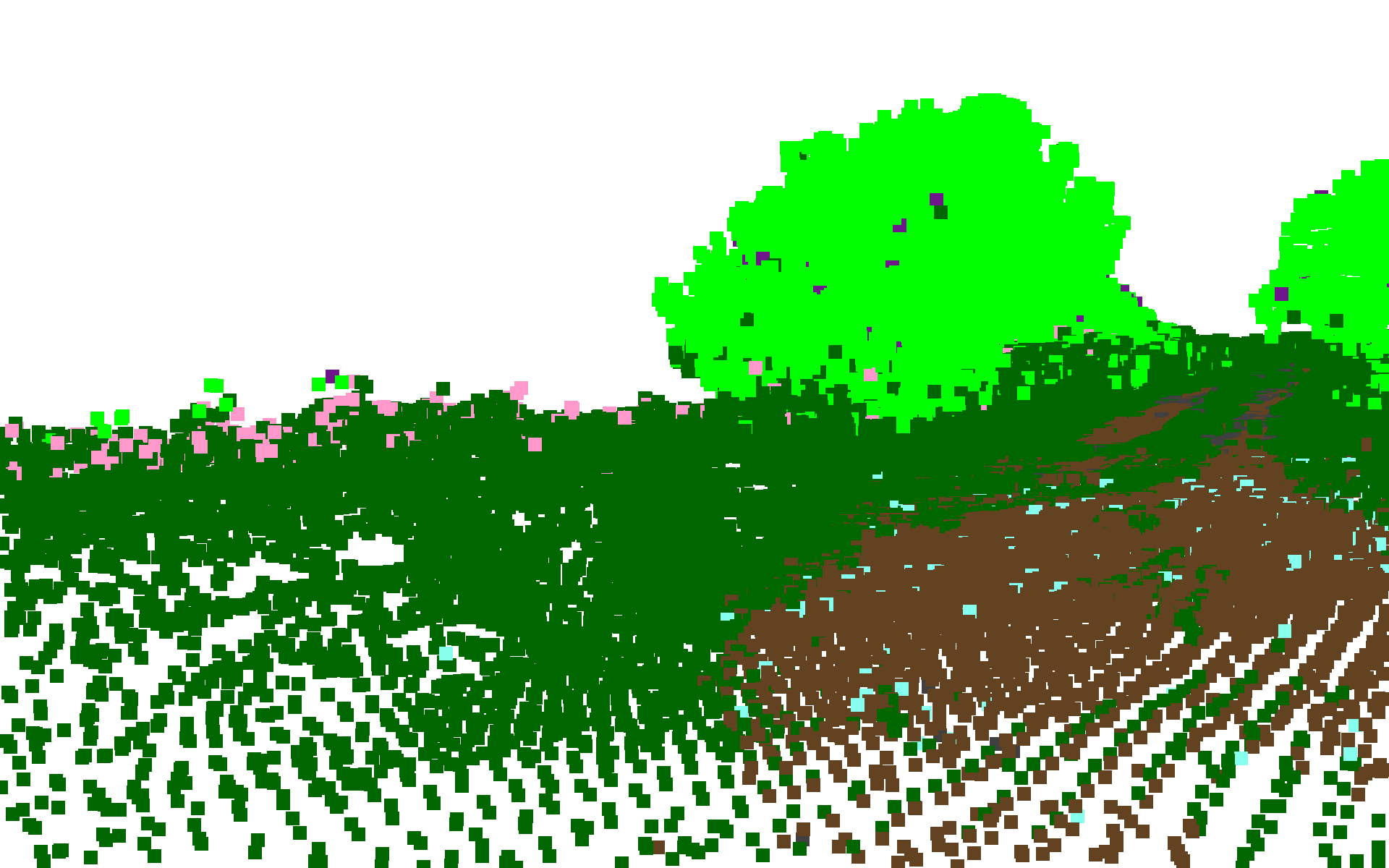} & 
        \includegraphics[width=0.95\linewidth]{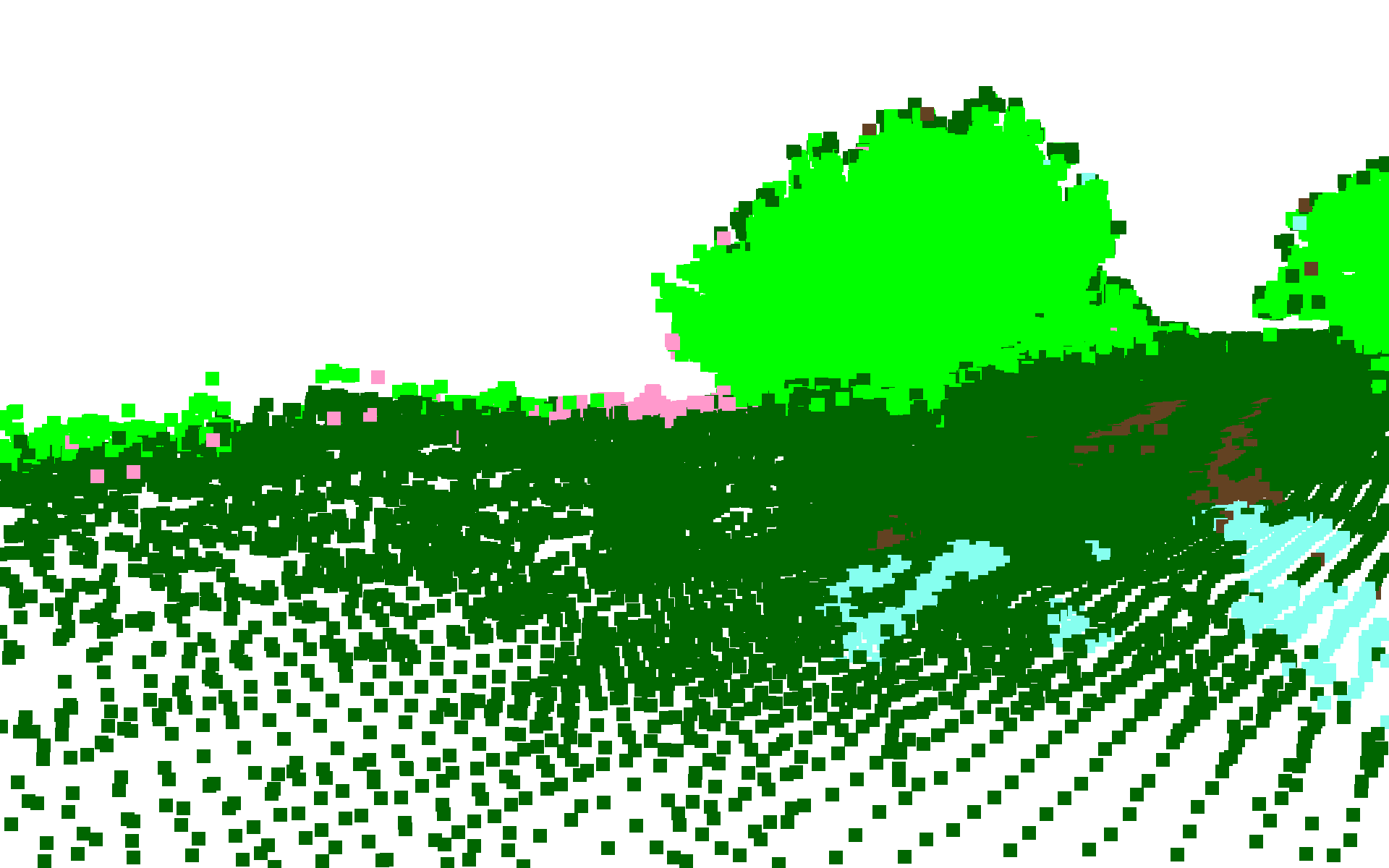} & 
        \includegraphics[width=0.95\linewidth]{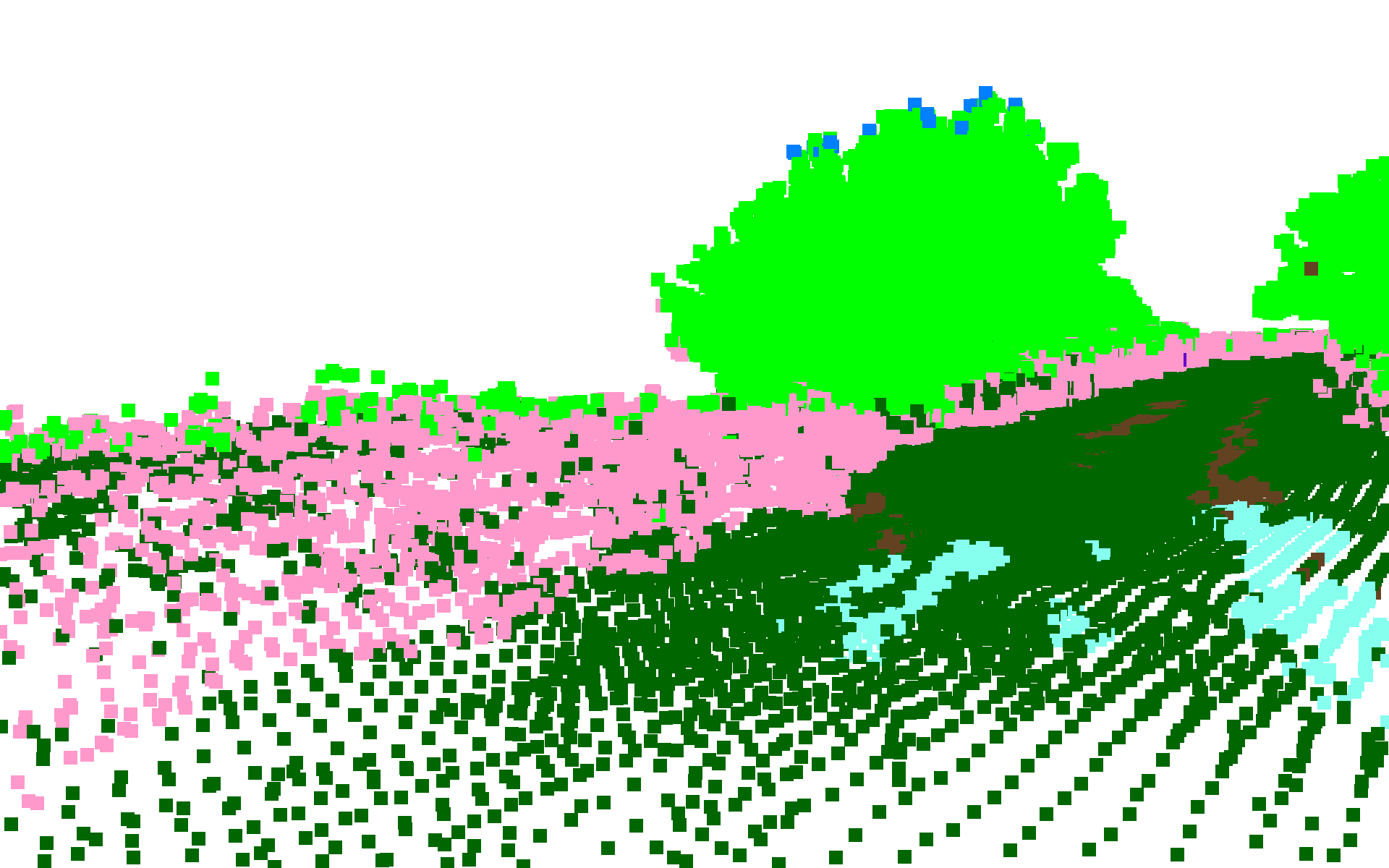} \\[0.4cm] 
        
        \midrule
        \multicolumn{5}{c}{\textbf{WildScenes}} \\
        \midrule
        
        \includegraphics[width=0.95\linewidth]{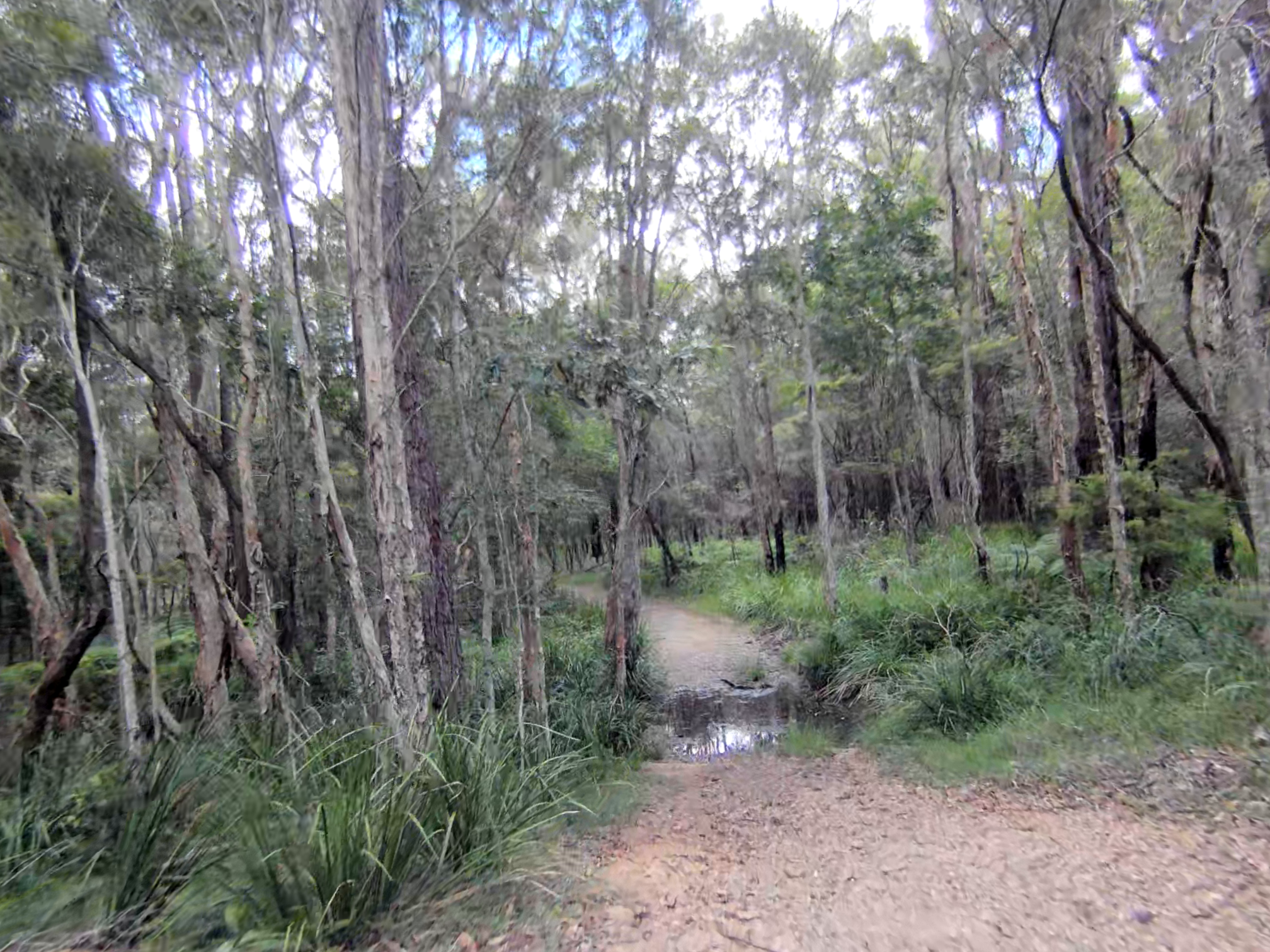} & 
        \includegraphics[width=0.95\linewidth]{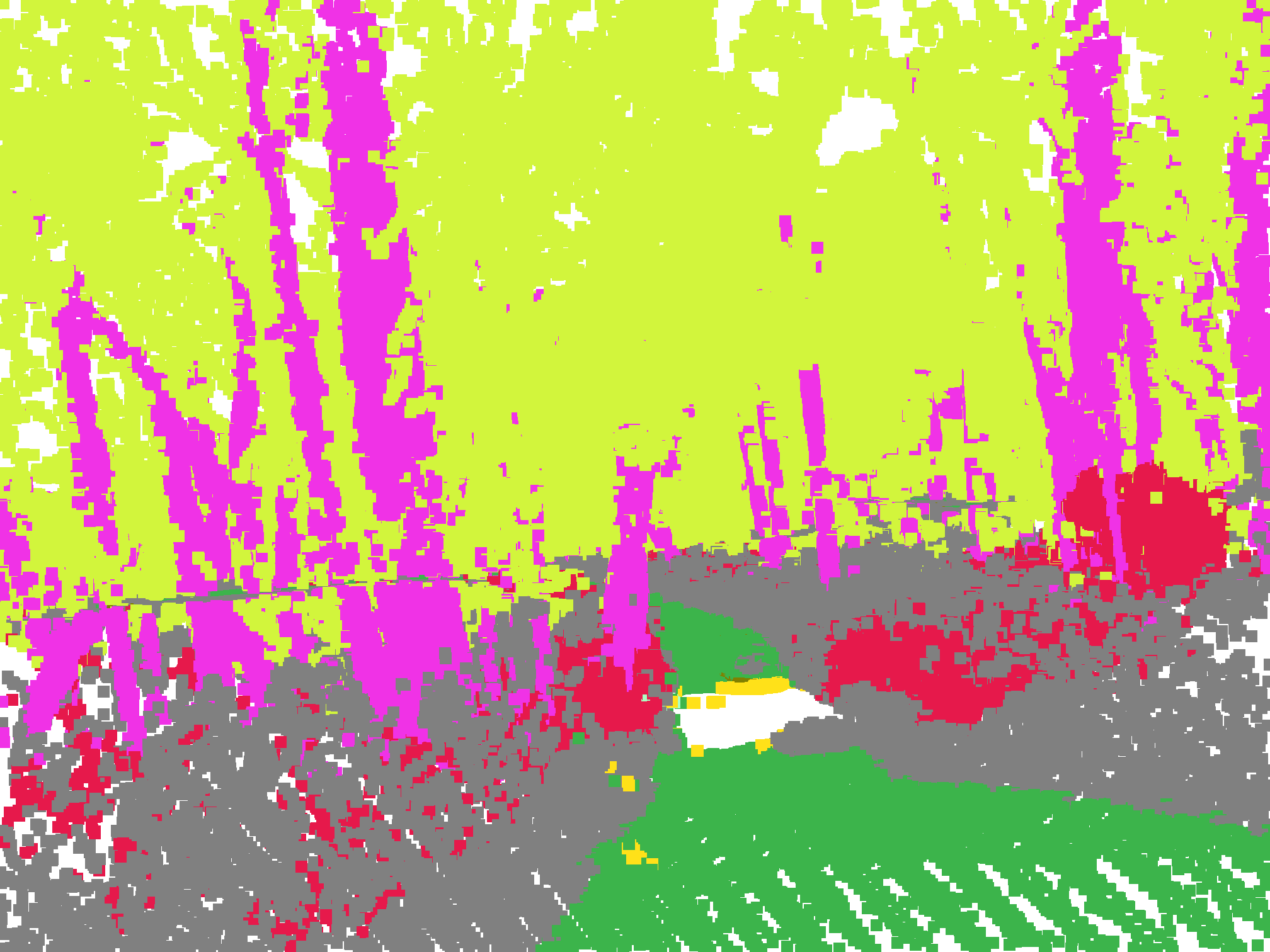} & 
        \includegraphics[width=0.95\linewidth]{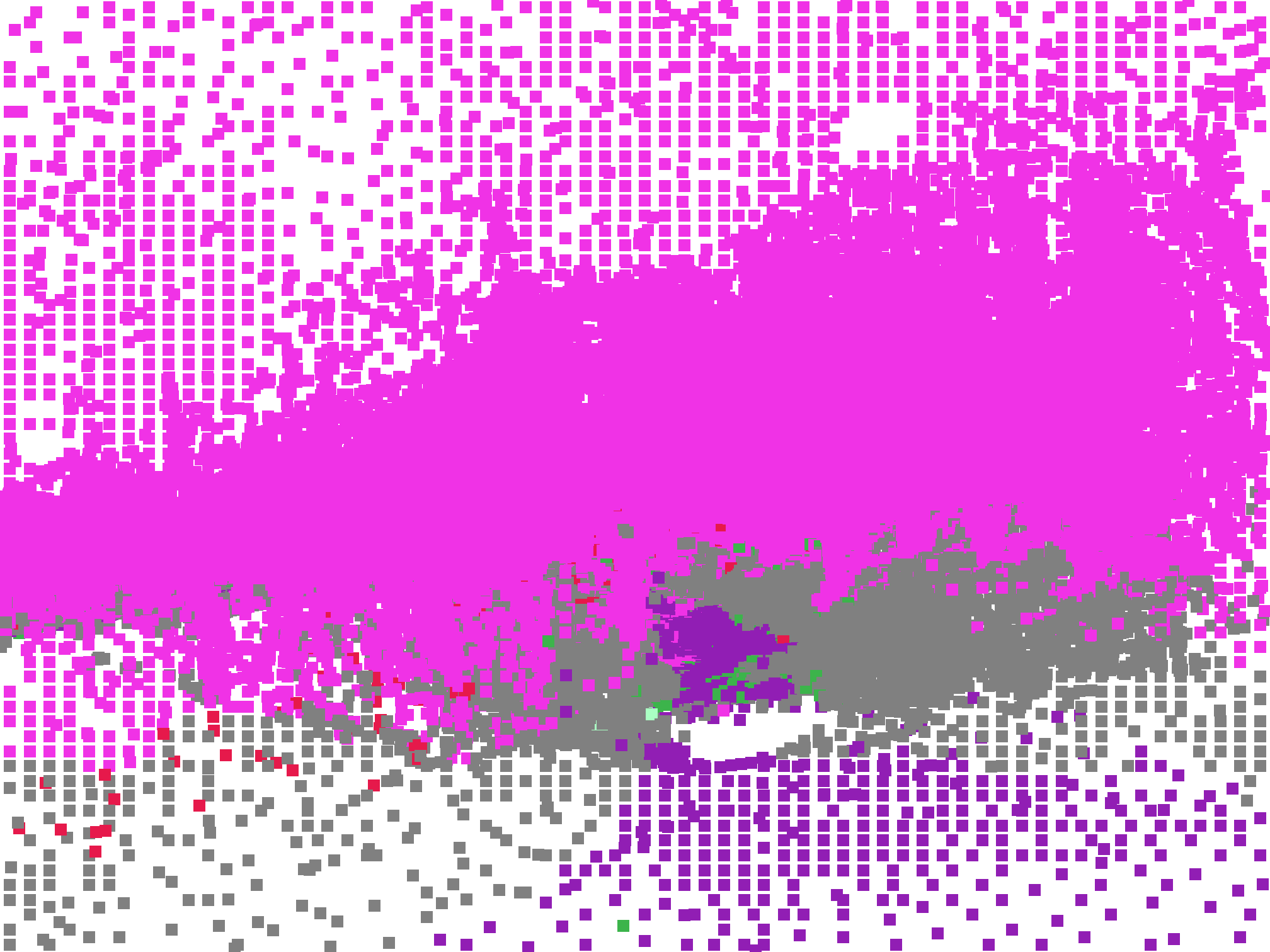} & 
        \includegraphics[width=0.95\linewidth]{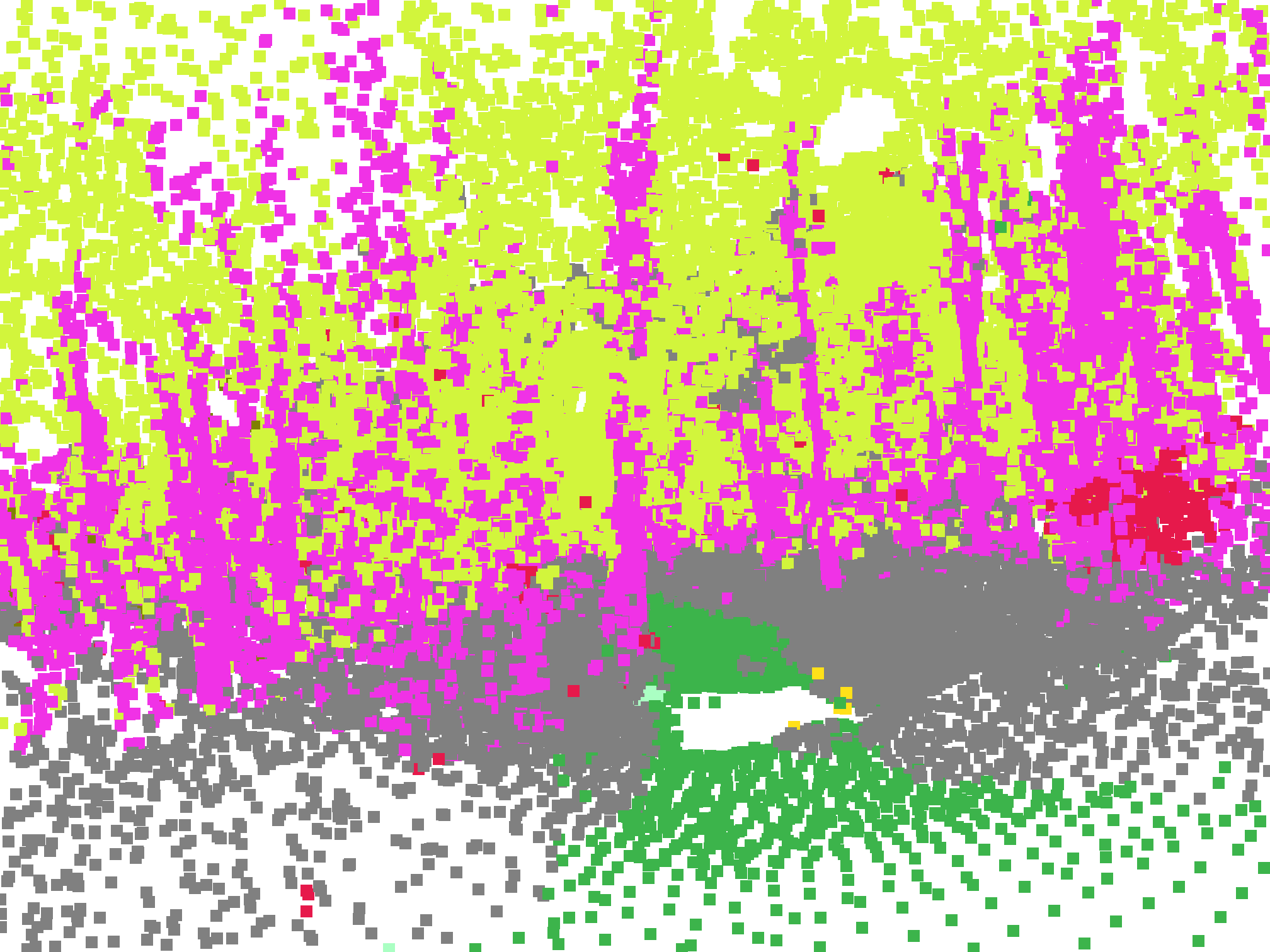} & 
        \includegraphics[width=0.95\linewidth]{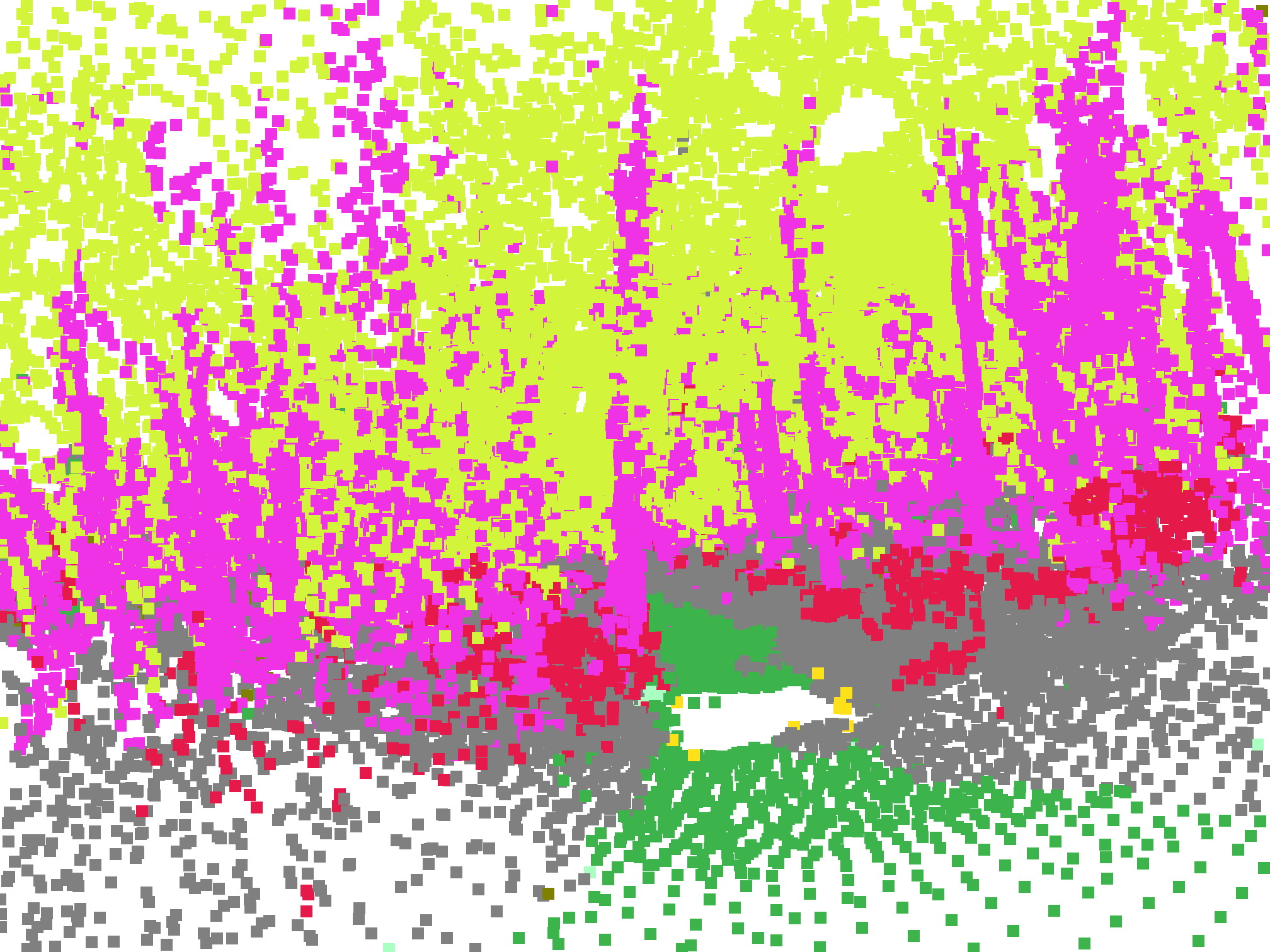} \\[0.2cm]
        
        \includegraphics[width=0.95\linewidth]{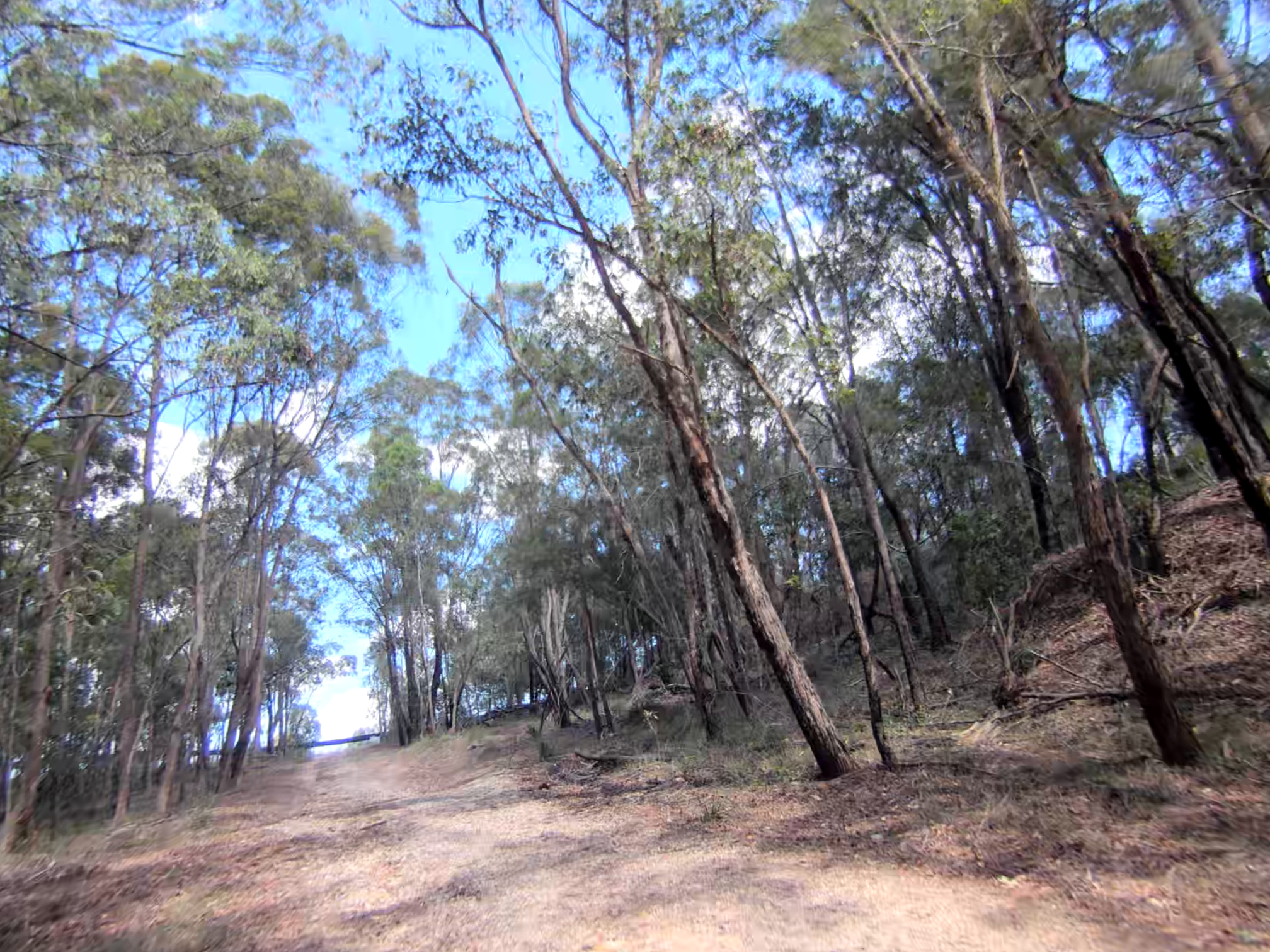} & 
        \includegraphics[width=0.95\linewidth]{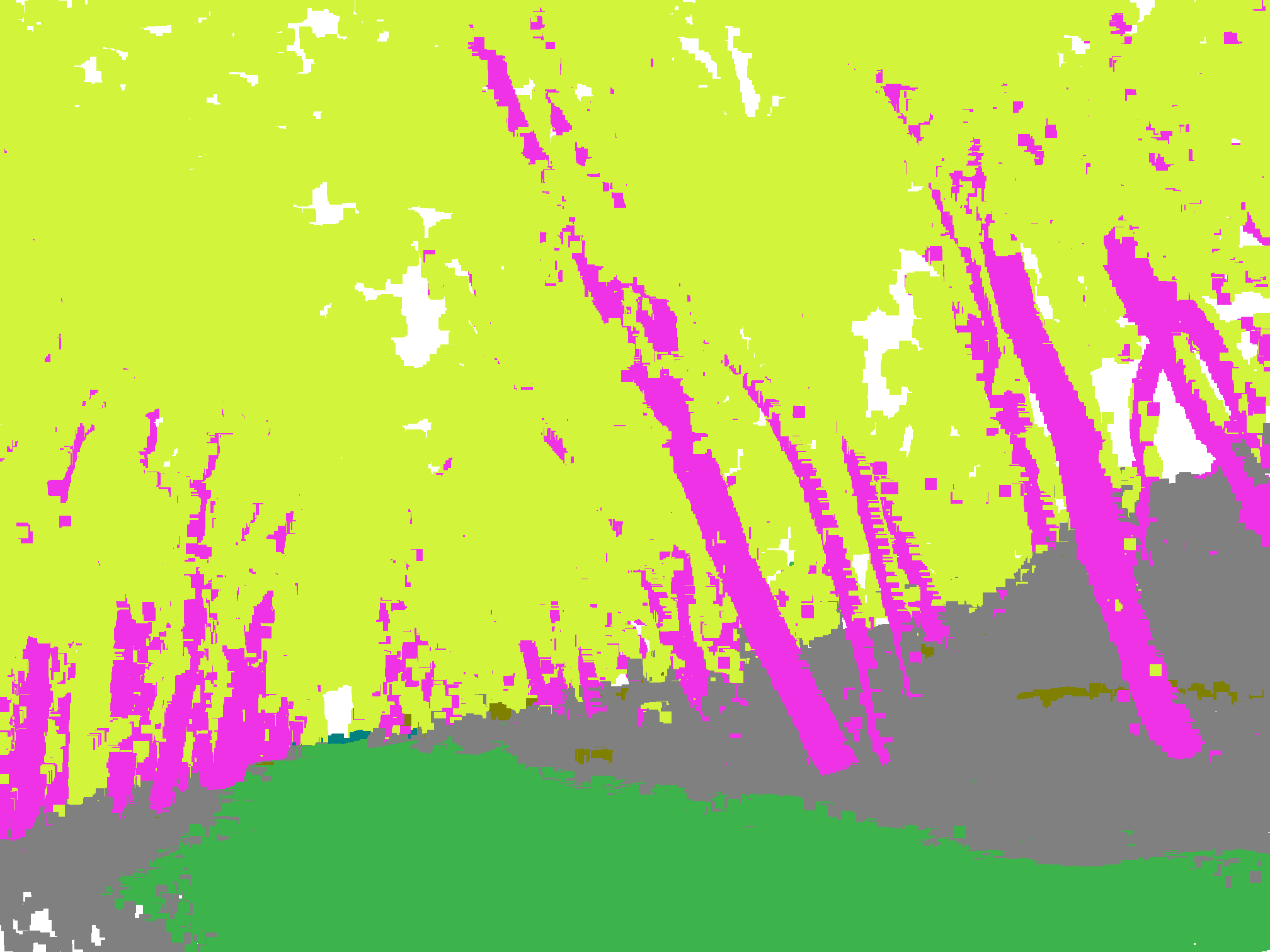} & 
        \includegraphics[width=0.95\linewidth]{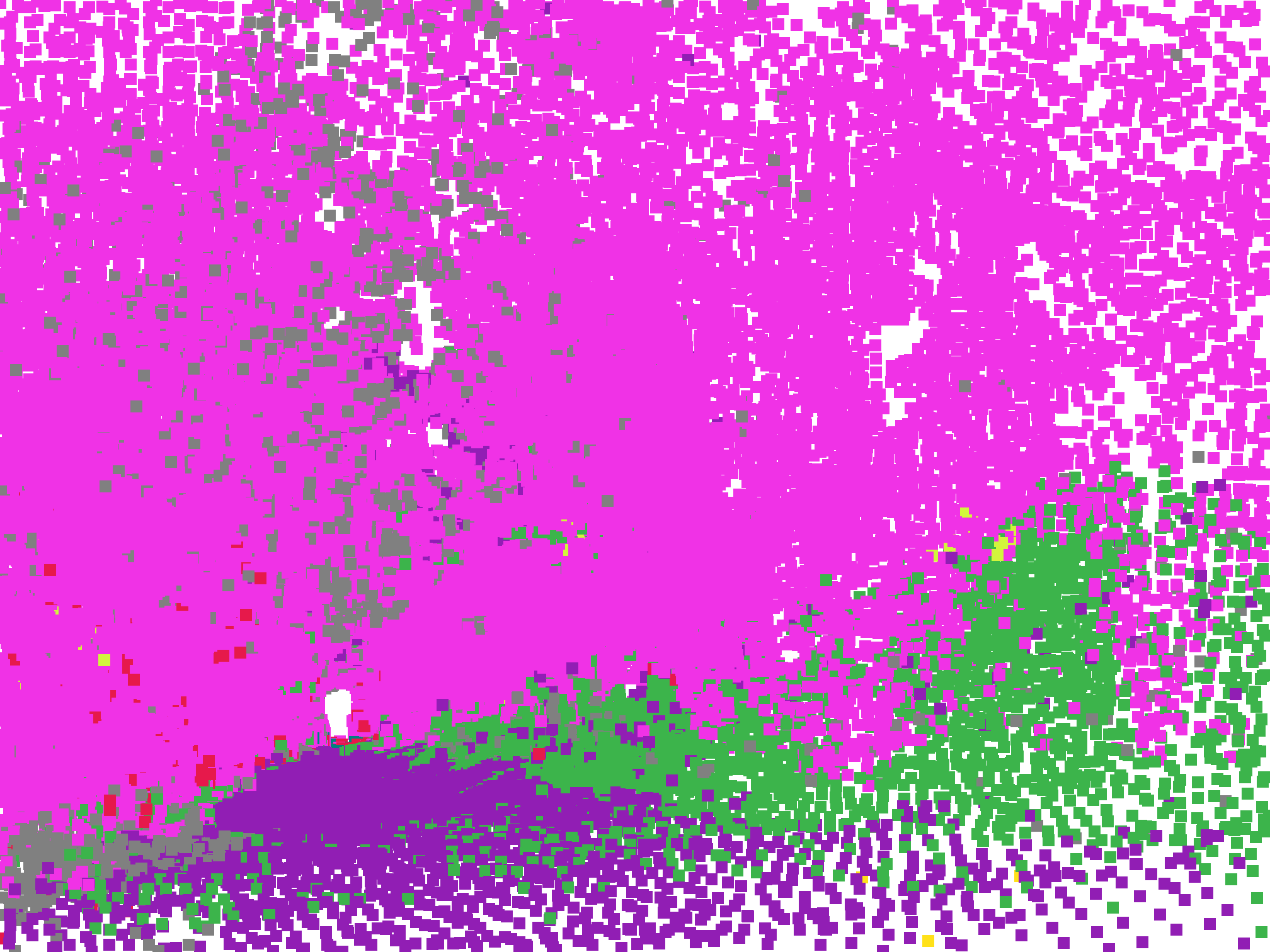} & 
        \includegraphics[width=0.95\linewidth]{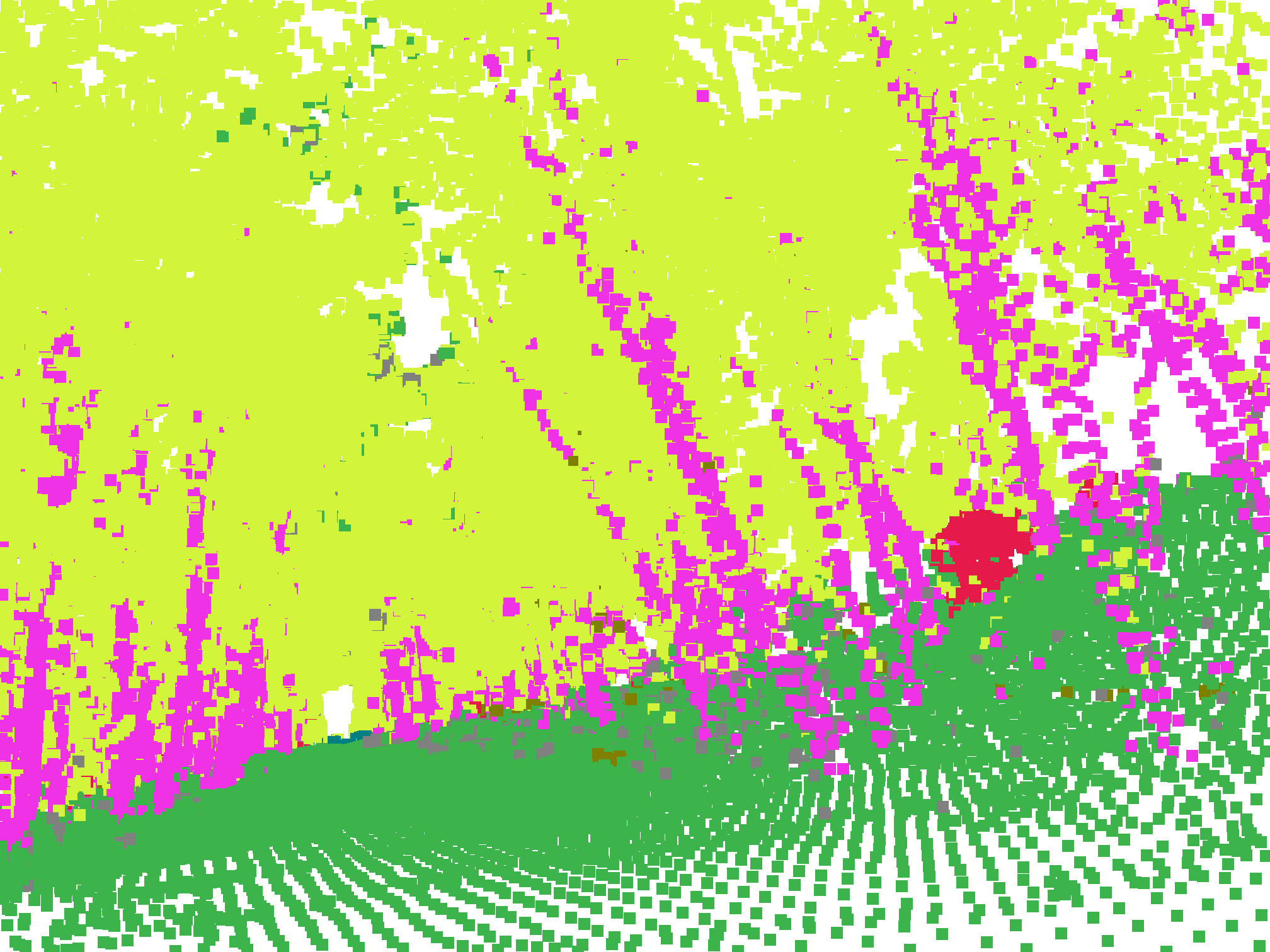} & 
        \includegraphics[width=0.95\linewidth]{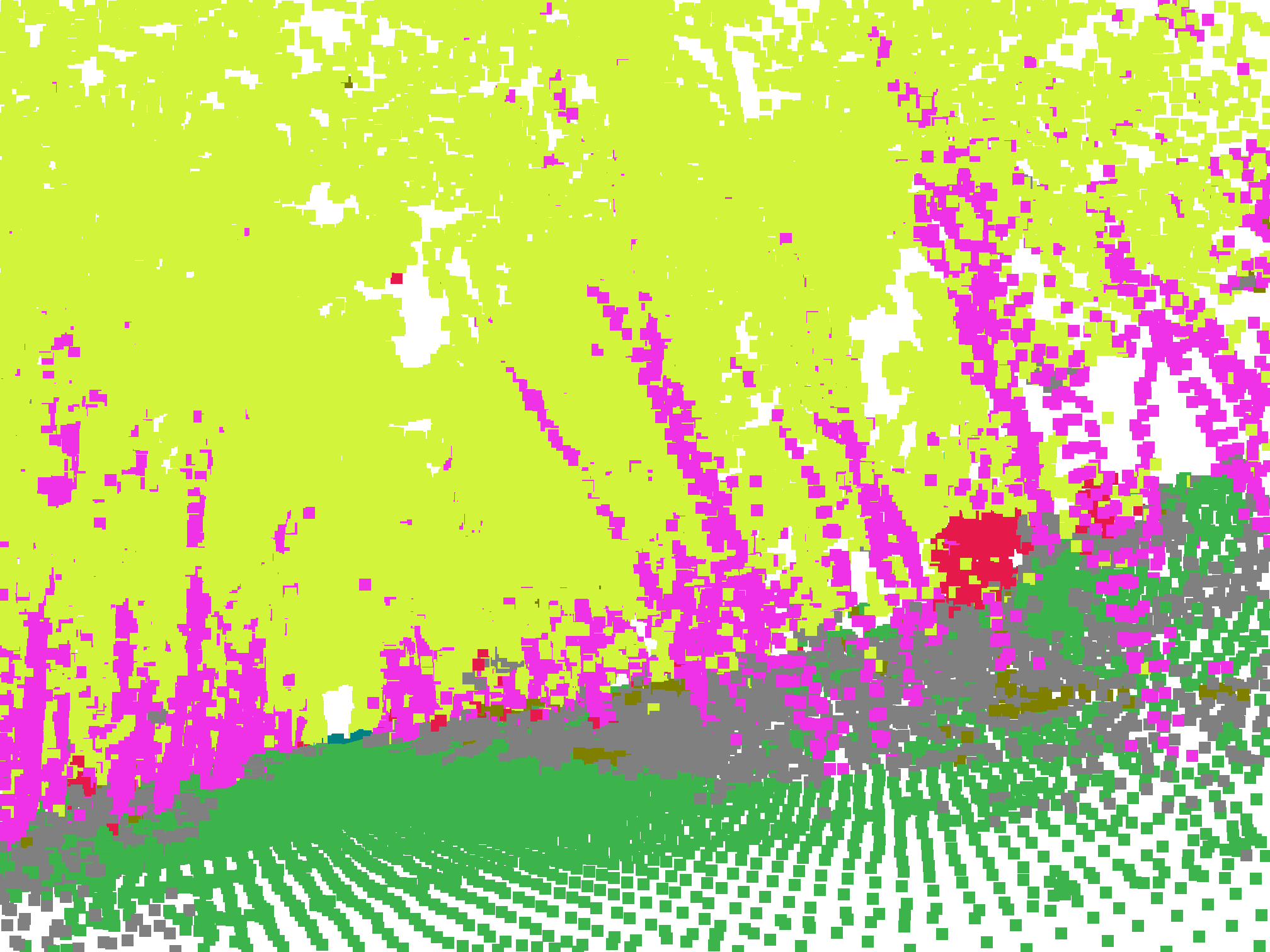} \\[0.2cm]

        \bottomrule
    \end{tabularx}
    \label{fig:qualitative}
\end{table*}

\begin{table*}[]
    \centering
    \caption{Examples of semantic pointcloud map predictions from birds eye view.}
    \begin{tabularx}{\textwidth}{*{4}{>{\centering\arraybackslash}X}}
        \toprule
        \textbf{Ground Truth} & \textbf{OTAS} & \textbf{SAM3} & \textbf{SAM3+OURS} \\
        \midrule
        
        \multicolumn{4}{c}{\textbf{RELLIS-3D}} \\
        \midrule
        
        \includegraphics[trim=0cm 10cm 0cm 10cm, clip, width=0.95\linewidth]{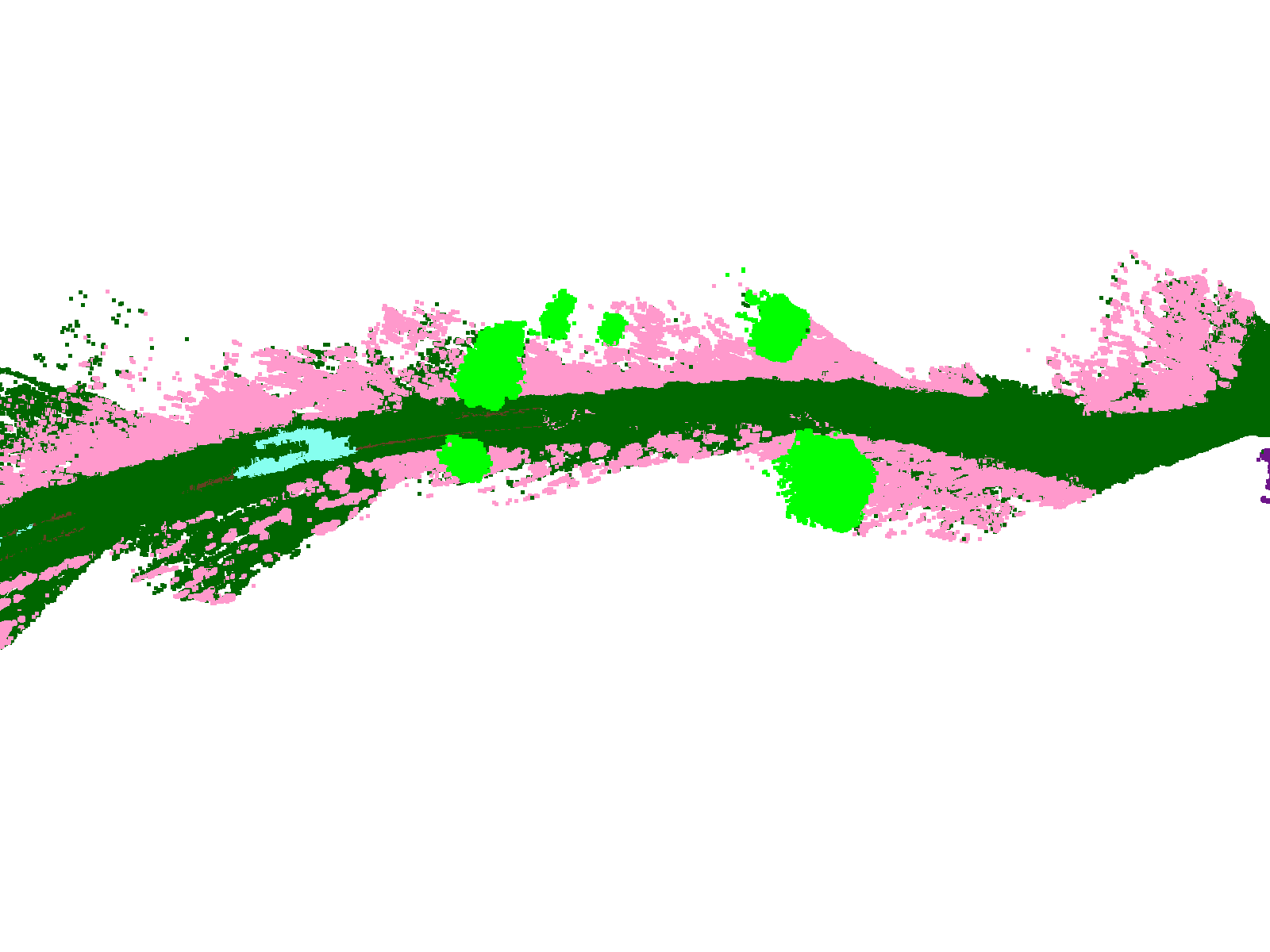} & 
        \includegraphics[trim=0cm 10cm 0cm 10cm, clip, width=0.95\linewidth]{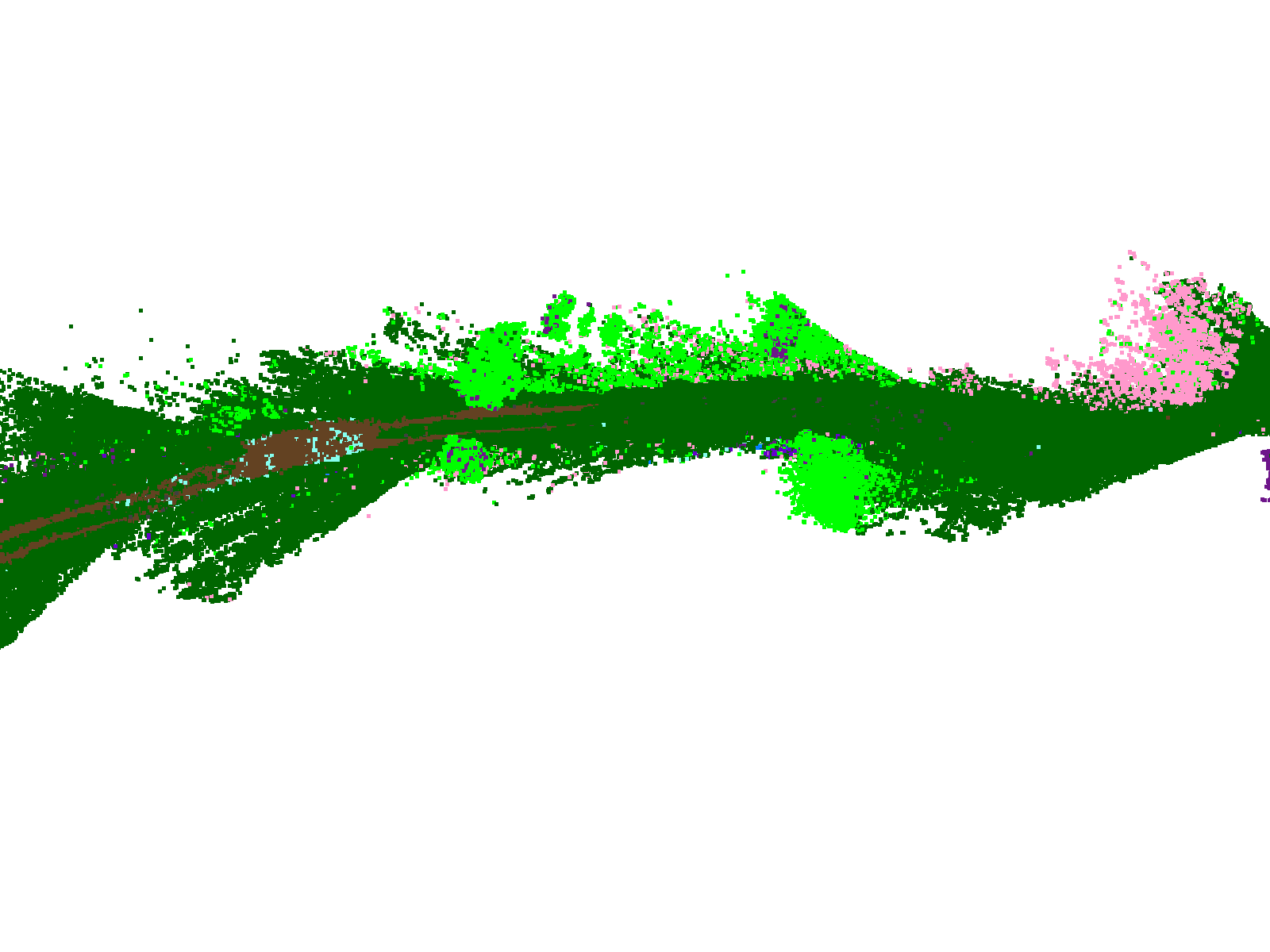} & 
        \includegraphics[trim=0cm 10cm 0cm 10cm, clip, width=0.95\linewidth]{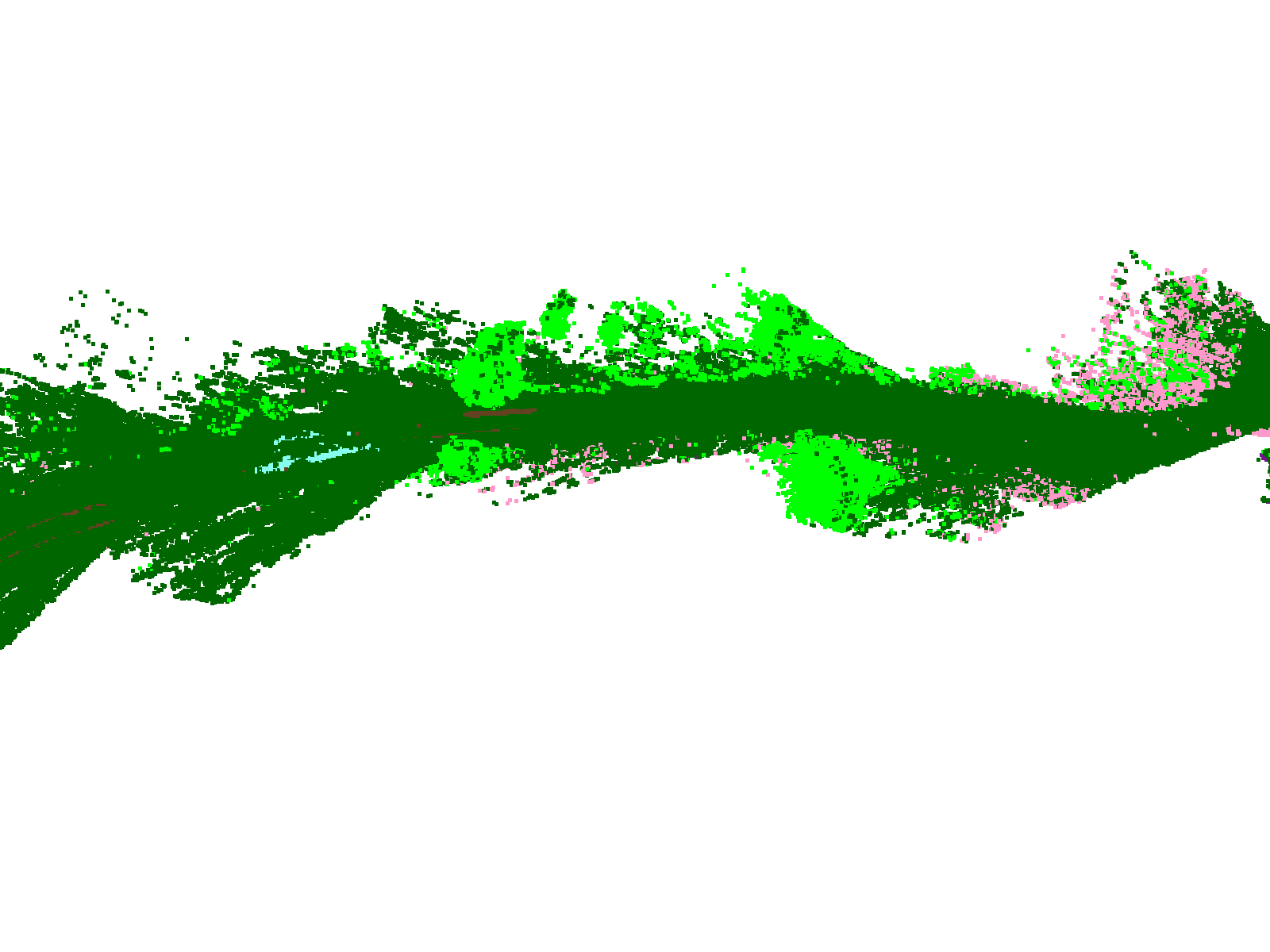} & 
        \includegraphics[trim=0cm 10cm 0cm 10cm, clip, width=0.95\linewidth]{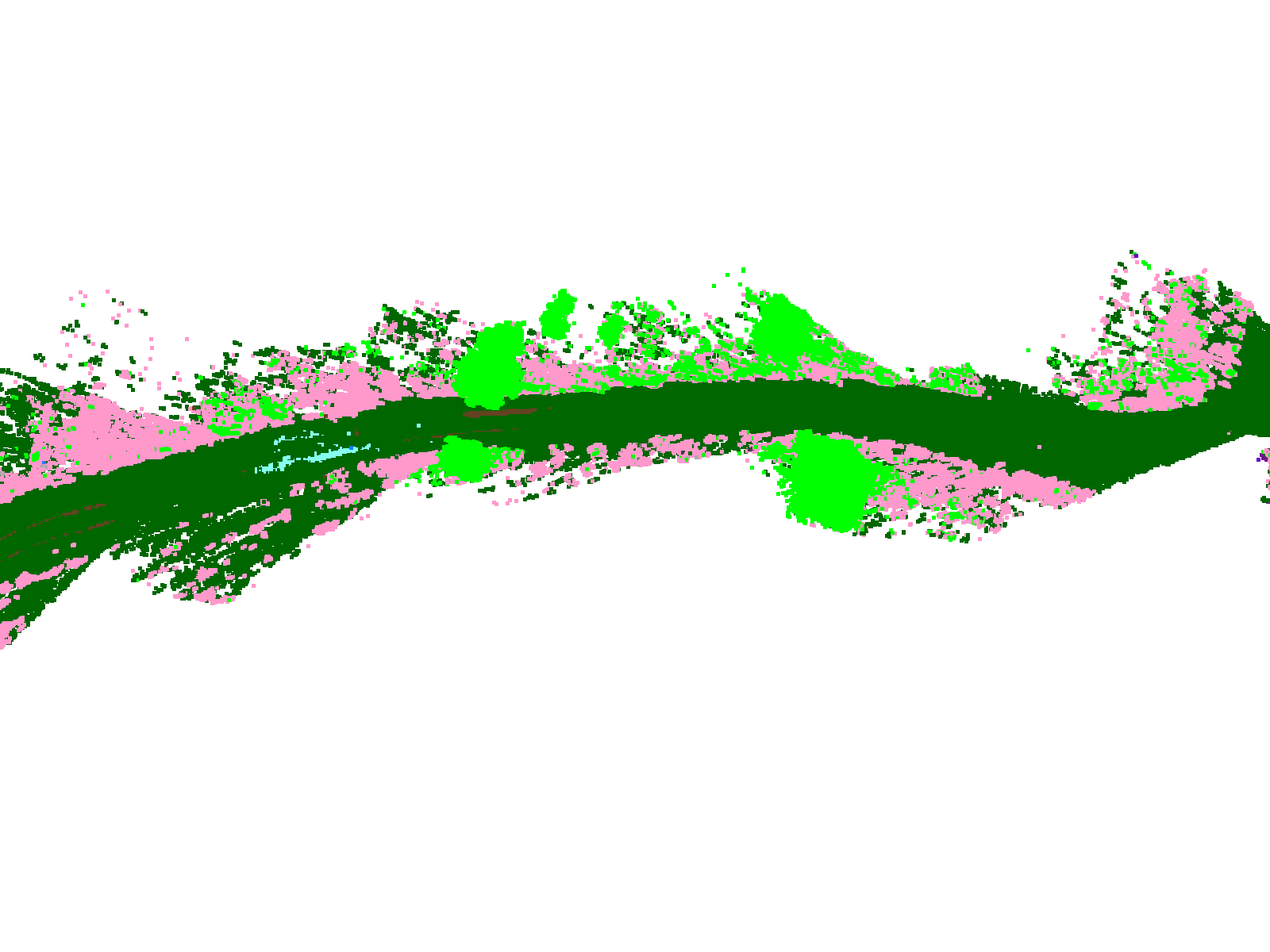}\\ 
        
        \midrule
        \multicolumn{4}{c}{\textbf{WildScenes}} \\
        \midrule

        \includegraphics[width=0.95\linewidth]{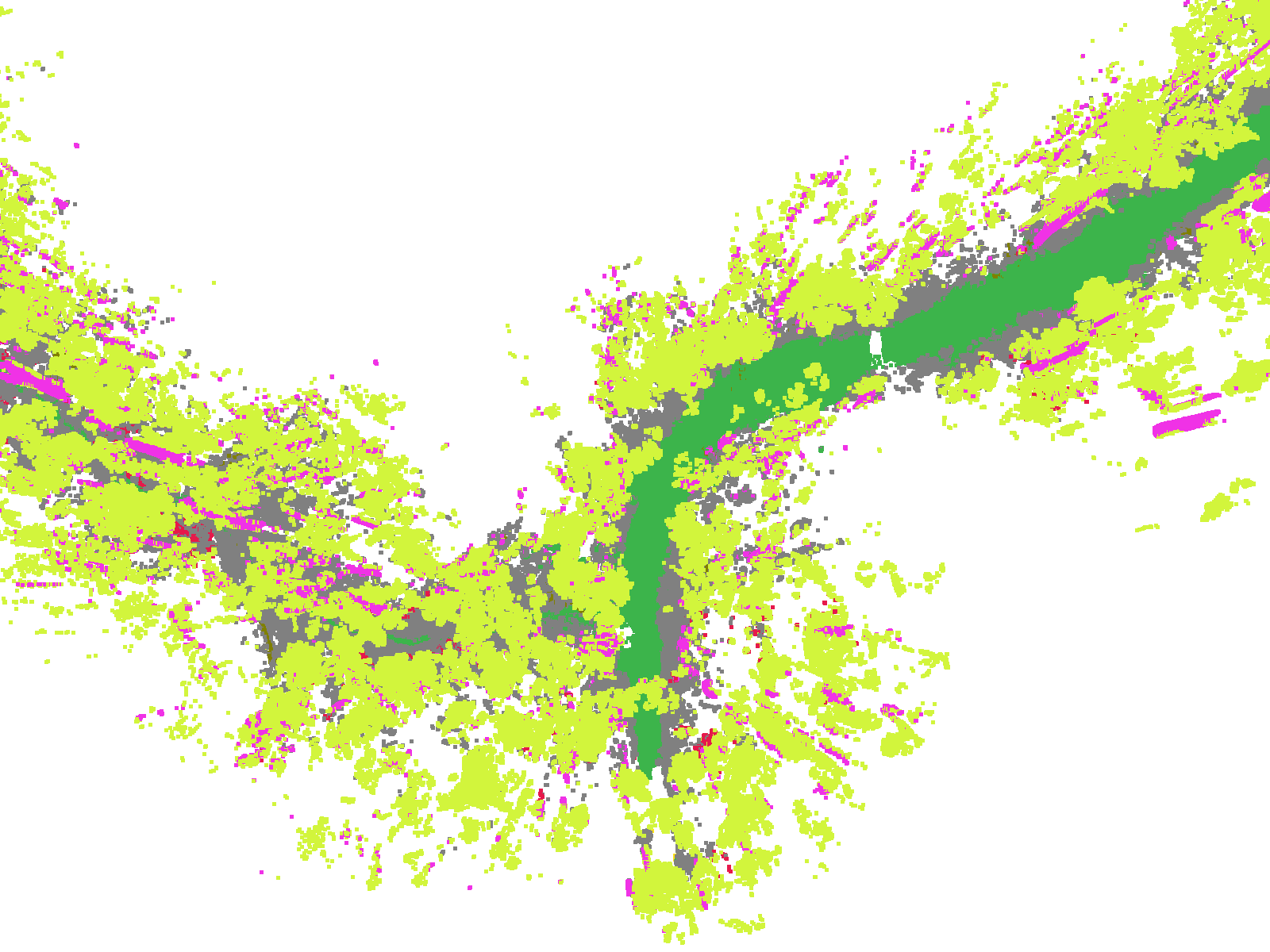} & 
        \includegraphics[width=0.95\linewidth]{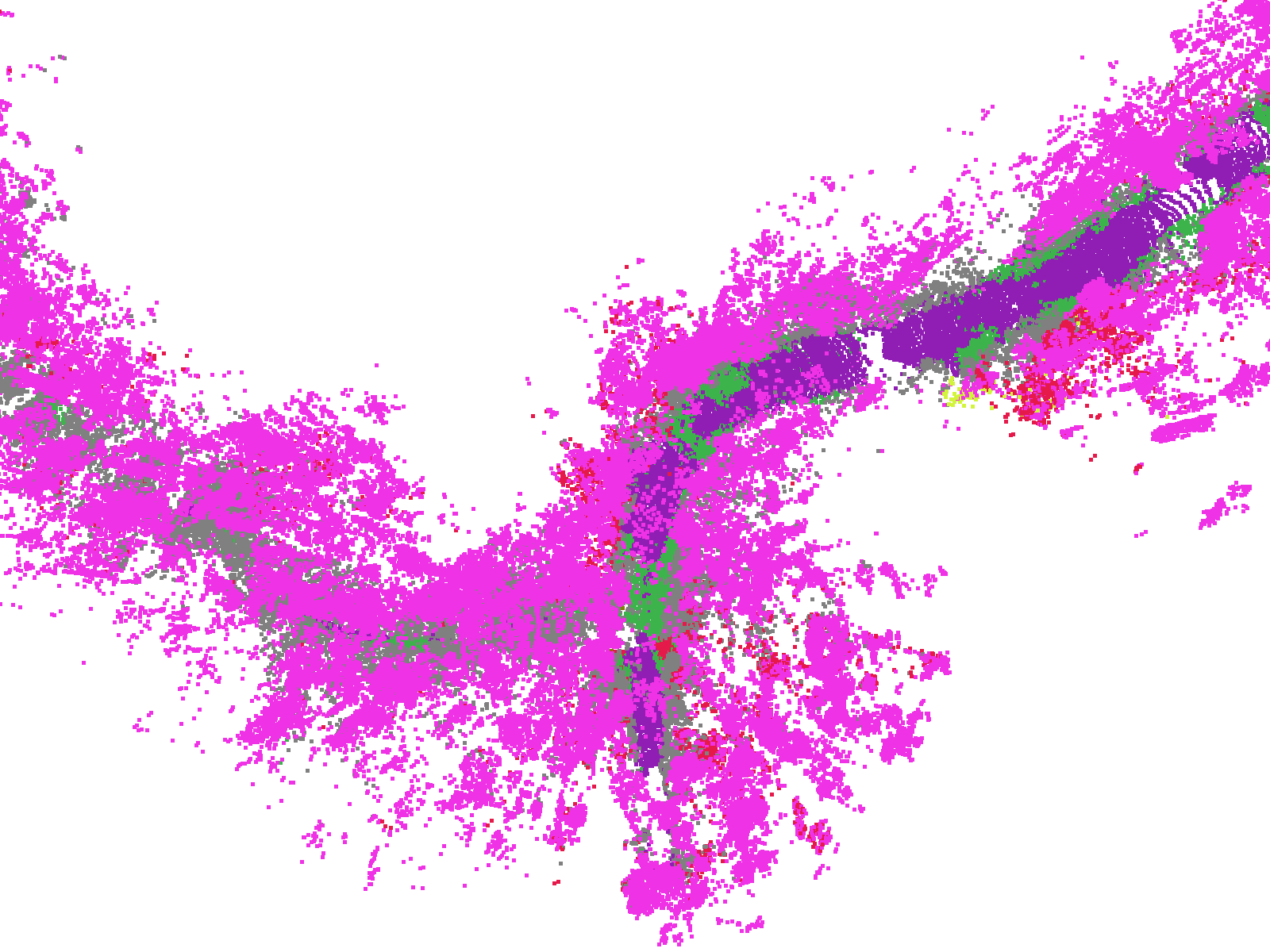} & 
        \includegraphics[width=0.95\linewidth]{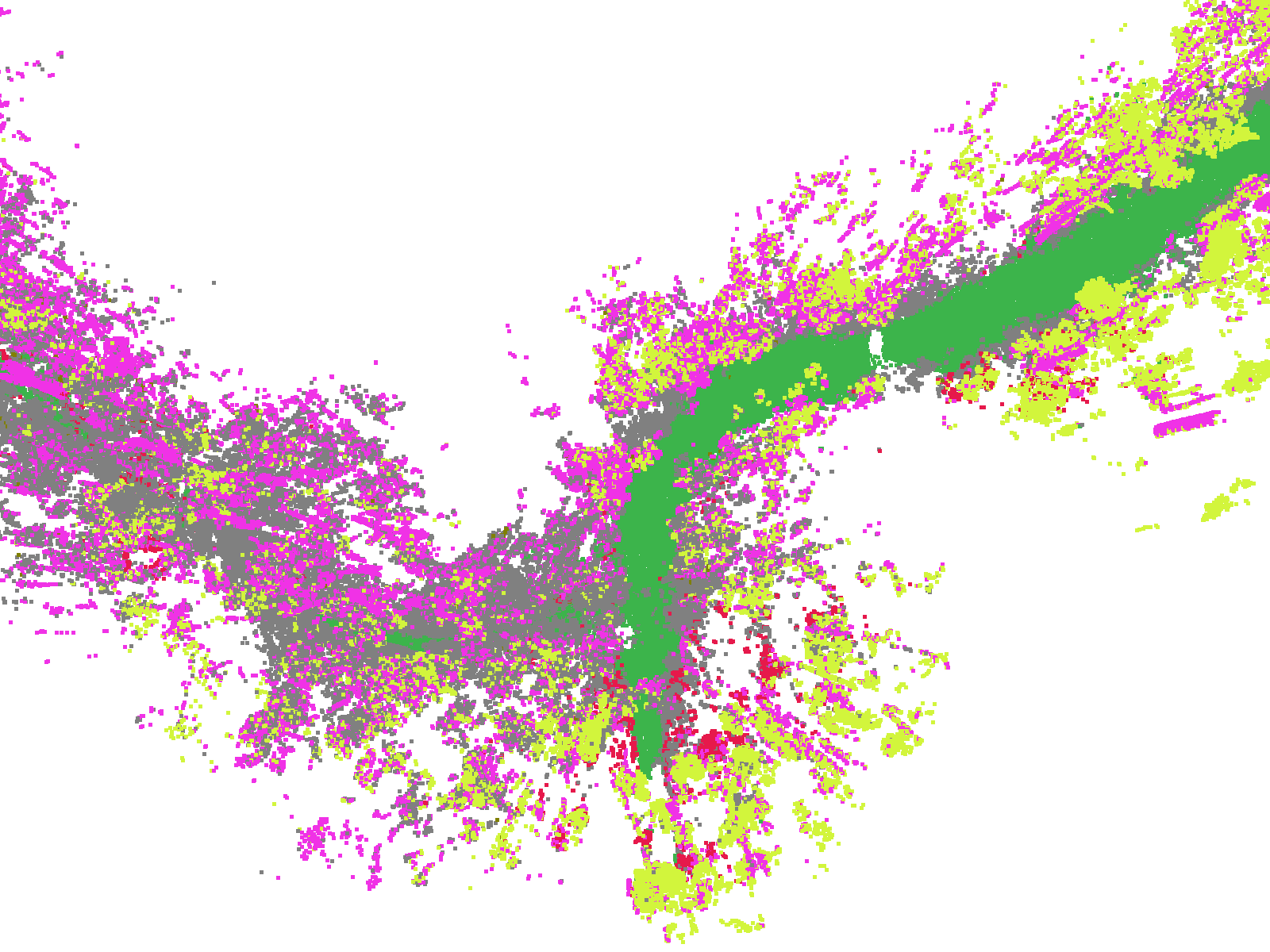} & 
        \includegraphics[width=0.95\linewidth]{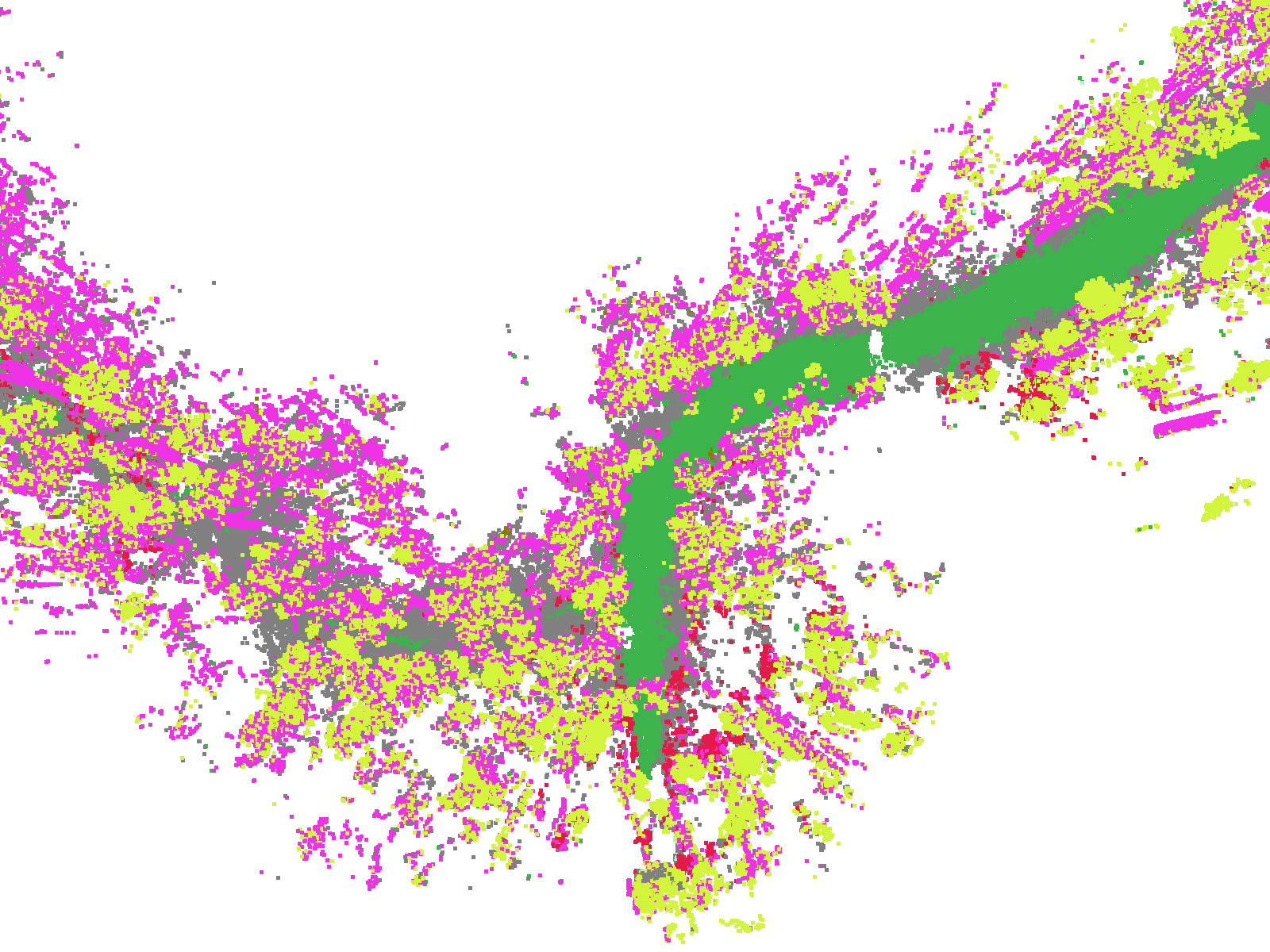} \\
        
        \bottomrule
    \end{tabularx}
    \label{fig:qualitative2}
\end{table*}

\begin{table}[]
    \centering
    \caption{Geometry prior ablations with different smoothing values $w$ for the geometric prior. ELEV defines the geometric prior based on elevation only, SPHE based on sphericity only and BOTH utilizes their joint distribution as presented in Section \ref{sec:fit}. Highest values \bf{bolded}.}
    \begin{tabular}{|l | c | c | c | c |}
        \hline
        & \multicolumn{2}{c|}{Wildscenes} & \multicolumn{2}{c|}{Rellis3d} \\
        \cline{2-5}
        Method & mIoU & Acc & mIoU & Acc \\
        \hline
        SAM 3 & 47.9 & 80.5 & 28.1 & 73.9 \\
         + SPHE $w$=0.1 & 48.9 & 81.5 & 31.4 & 78.3 \\
         + SPHE $w$=0.25 & 49.2 & 81.6 & 32.4 & 81.8 \\
         + SPHE $w$=0.5 & 47.8 & 80.2 & 32.5 & 82.5 \\
         + SPHE $w$=1.0 & 43.5 & 77.1 & 30.8 & 80.3 \\
         + ELEV $w$=0.1 & 49.0 & 81.6 & 30.8 & 77.4 \\
         + ELEV $w$=0.25 & 48.2 & 81.3 & 33.0 & 81.2 \\
         + ELEV $w$=0.5 & 45.8 & 79.9 & 34.5 & 82.6 \\
         + ELEV $w$=1.0 & 38.8 & 76.7 & 34.2 & 82.8 \\
         + BOTH $w$=0.1 & \textbf{49.6} & 82.4 & 32.4 & 78.7 \\
         + BOTH $w$=0.25 & \textbf{49.6} & \textbf{82.6} & 34.4 & 82.8 \\
         + BOTH $w$=0.5 & 47.4 & 81.6 & \textbf{35.5} & \textbf{83.9} \\
         + BOTH $w$=1.0 & 37.8 & 79.1 & 33.7 & 82.6 \\
        \hline
         DINOv2.txt & 21.3 & 24.0 & 21.0 & 65.5 \\
         + BOTH $w$=0.1 & 22.4 & 28.3 & 21.8 & 66.8 \\
         + BOTH $w$=0.25 & \textbf{23.4} & 33.2 & 22.5 & 68.0 \\
         + BOTH $w$=0.5 & 23.0 & 38.1 & \textbf{23.1} & \textbf{69.0} \\
         + BOTH $w$=1.0 & 19.3 & \textbf{42.8} & 22.9 & 68.4 \\
        \hline
        OpenSeg & 27.0 & 73.9 & 21.5 & 75.1 \\
         + BOTH $w$=0.1 & 27.3 & 74.1 & 22.1 & 76.3 \\
         + BOTH $w$=0.25 & 27.9 & 74.4 & 23.1 & 77.4 \\
         + BOTH $w$=0.5 & 28.8 & 74.6 & 24.0 & 78.5 \\
         + BOTH $w$=1.0 & \textbf{29.1} & \textbf{75.0} & \textbf{24.5} & \textbf{79.6} \\
        \hline
    \end{tabular}
    \label{tab:ablations}
\end{table}

\section{Discussion}
\label{sec:conclusion}
This paper introduced a method to improve zero-shot segmentation in offroad scenes with geometric priors computed from the 3D data. The geometric priors increased performance especially on natural classes like bush, grass, dirt, water and puddle while performance on structured classes was very little affected. Our explanation for this behavior is two-fold. First, the zero-shot vision foundation models are more prone to make mistakes on natural classes due to the limited training data from these domains, leaving more room for improvement. Second, the elevation-sphericity histogram has better separation between natural classes than between structured ones. In Figure \ref{fig:rellis_hist},  barrier, fence and pole all share quite similar distributions while grass, bush and tree are more clearly separated. 

While deep learning models can extract more complex features from 3D pointclouds, the proposed approach has a few main advantages over them. Deep learning models require training on each new environment while the distribution fitting process is much faster and requires less computation. Additionally, the proposed approach is more transparent: the geometric separation between classes can be directly evaluated from the fitted histograms. Finally, for many natural classes elevation and sphericity provide sufficient separation when fused with the vision foundation model predictions as the 3D shapes are simpler compared to cluttered indoor scenes. 

The performance of our method is limited by class separability within the histogram; consequently, classes sharing similar elevation and sphericity values cannot be distinguished. Furthermore, our approach relies on the vision foundation model to provide confident detections for each class in order to fit these distributions. Any bias or error on the vision foundation models detections affects the fitted histogram and consequentially the computed priors. In future work, more geometric features could be added to achieve more discriminative prior distributions or alternatively pre-trained point-encoders could be used to extract more complex point features. Another future challenge is calibrating the vision foundation model and geometric prior confidences for improved fusion and uncertainty estimation.   

While histograms can capture arbitrary distribution shapes, they also introduce significant limitations. Specifically, a query bin's prior depends solely on its own value, without accounting for neighboring bins. Consequently, an empty bin adjacent to high-density regions is treated identically to an empty bin in a sparse region. Future work could address this limitation by applying spatial smoothing to the histograms.  Parameterized distributions, such as 2D Gaussians are also a potential solution but they also restrict the shape of the distribution.    

\section*{ACKNOWLEDGMENTS}
Acknowledgments that can be used to identify the authors have been excluded from this version.  
Large language models (Claude 5 Sonnet and Opus) were
used in the code implementation of the presented methodology,
and experimental evaluation (Sections \ref{sec:methods} and \ref{sec:results}).
The authors reviewed all AI-assisted content and take full
responsibility for it.

\bibliographystyle{IEEEtran}
\bibliography{root} 
	
\end{document}